%% file: main.tex
\documentclass{article}

\PassOptionsToPackage{square,numbers}{natbib}
\usepackage[final]{neurips_2022}

\usepackage[T1]{fontenc}    
\usepackage{hyperref}       
\usepackage{url}            
\usepackage{booktabs}       
\usepackage{amsfonts}       
\usepackage{nicefrac}       
\usepackage{microtype}      
\usepackage{xcolor}         
\usepackage{import}
\usepackage{amsmath}
\usepackage{authblk}
\usepackage{amssymb}
\usepackage{comment}
\usepackage[]{graphicx}
\usepackage{makecell}
\usepackage{colortbl}
\usepackage{subcaption}
\usepackage[rightcaption]{sidecap} 
\usepackage[outputdir=latexmk,finalizecache=false, frozencache=true]{minted} 
\usepackage{tikz}
\usepackage[ruled]{algorithm2e} 

\title{Okapi: Generalising Better by \\ Making Statistical Matches Match}

\author{%
  \textbf{Myles Bartlett}$^{1}$\thanks{Corresponding author: \texttt{m.bartlett@sussex.ac.uk}.} \quad
  \textbf{Sara Romiti}$^{1}$ \quad 
  \textbf{Viktoriia Sharmanska}$^{1,2}$ \quad 
  \textbf{Novi Quadrianto}$^{1, 3, 4}$ \\
  $^1$Predictive Analytics Lab, University of Sussex \quad  
  $^2$Imperial College London \\
  $^3$BCAM Severo Ochoa Strategic Lab on Trustworthy Machine Learning \\
  $^4$Monash University, Indonesia 
}

\import{./}{commands.tex}

\begin{document}

\maketitle

\begin{abstract}
\import{./}{abstract.tex}
\end{abstract}

\section{Introduction}\label{sec:intro}
\import{./}{introduction.tex}

\section{Preliminaries}\label{sec:prelims}
\import{./}{preliminaries.tex}


\section{Method}\label{sec:method}
\import{./}{methodology.tex}

\section{Related Work}\label{sec:related_work}
\import{./}{related_work.tex}

\section{Experiments}\label{sec:exps}
\import{./}{experiments.tex}

\section{Conclusion}\label{sec:conclusion}
\import{./}{conclusion.tex}


\bibliography{bibfile.bib}

\newpage
\appendix
\import{./}{supplementary.tex}


\end{document}

%% file: commands.tex
\newcommand{\D}{\mathcal{D}}
\newcommand{\M}{\mathcal{M}}
\newcommand{\NM}{N_\mathcal{M}}
\newcommand{\CNN}{CaliperNN}
\newcommand{\mr}{\mathrm}
\newcommand{\mc}{\mathcal}
\newcommand{\mbf}{\mathbf}

\definecolor{RHL}{rgb}{191, 255, 105}

\definecolor{oecol}{rgb}{1, 0.3, 0}
\definecolor{tecol}{rgb}{0.05, 0.35, 0.85}

%% file: abstract.tex
We propose \emph{Okapi}, a simple, efficient, and general method for robust semi-supervised
learning based on online statistical matching. 
Our method uses a nearest-neighbours-based matching procedure to generate cross-domain views for a
consistency loss, while eliminating statistical outliers.
In order to perform the online matching in a runtime- and memory-efficient way, we draw upon the
self-supervised literature and combine a memory bank with a slow-moving momentum encoder.
The consistency loss is applied within the feature space, rather than on the predictive
distribution, making the method agnostic to both the \emph{modality} and the \emph{task} in
question.
We experiment on the WILDS 2.0 datasets \citep{SagWeiLeeGaoetal22}, which significantly expands the
range of modalities, applications, and shifts available for studying and benchmarking real-world
unsupervised adaptation. 
Contrary to \cite{SagWeiLeeGaoetal22}, we show that it is in fact possible to leverage additional
unlabelled data to improve upon empirical risk minimisation (ERM) results with the right method.
Our method outperforms the baseline methods in terms of out-of-distribution (OOD) generalisation on
the iWildCam (a multi-class classification task) and PovertyMap (a regression task) image datasets
as well as the CivilComments (a binary classification task) text dataset.
Furthermore, from a qualitative perspective, we show the matches obtained from the learned
encoder are strongly semantically related. 
Code for our paper is publicly available at ~\url{https://github.com/wearepal/okapi/}.

%% file: introduction.tex
Machine learning models have been deployed for safety-critical applications such as disease
diagnosis \citep{watson2019clinical} and self-driving cars \citep{yu2020bdd100k}, and in socially
important contexts such as the allocation of healthcare, education, and credit (e.g.
\cite{DunYiLanReetal19, HurAde17}). 
%
%
Many machine learning algorithms, however, rely on supervision from a large amount of labelled
data, and are typically trained to exploit complex relationships and distant correlations present
in the training dataset. 
This strategy has proven to be effective in the setting when we have training (source) and test
(target) data that are i.i.d.

In reality, machine learning models are often deployed on target data whose distribution is
different from the source distribution they were trained on. 
For example, in the task of classifying animal species in a camera trap image, one aims to learn a
model that can generalise to new camera trap locations despite variations in illumination,
background, and label frequencies, given training examples from a limited set of camera trap
locations.
Exploiting correlations that only hold in these limited locations but not in the new locations can
hurt out-of-distribution (OOD) generalisation.
While we only have a small subset of camera traps that have their images labelled, we have a large
amount of unlabelled data from the other camera traps that capture diverse operating conditions. 
In general, unlabelled data is much more readily available than labelled data and can often be
obtained from distributions beyond the source distribution.
Taking advantage of these unlabelled data during training is a key element to build robust models
that have good OOD performance without sacrificing in-distribution (ID) performance.
%

Our work is a direct response to the empirical conclusions of \cite{SagWeiLeeGaoetal22} for the
WILDS 2.0 dataset, which extends the WILDS benchmark datasets of \cite{koh2021wilds} through the
addition of unlabelled data. %
In \citet{SagWeiLeeGaoetal22} a variety of state-of-the-art methods leveraging unlabelled data, including
domain-invariant, self-training, and self-supervised methods were evaluated for their ability to
improve OOD generalisation.
In the all but a few cases, however, these methods failed to outperform the combination of effective
data-augmentation and standard empirical risk minimisation (ERM), and among those select cases
none persisted across datasets.

We show that it is possible to make effective use of large volumes of unlabelled data
as supplement to a smaller set of labelled data, from a limited set of domains, to achieve
strong generalisation to data from domains outside the training distribution.
%
We turn to a statistical matching (SM) framework \cite{RomInsShaQua22,rosenbaum1985constructing, rubin1973matching}, a model-based approach for providing joint information on variables and
indicators collected through multiple sources. 
SM has been widely utilised to assess the effect of interventions in numerous fields, such as
education, medical and community policies (e.g. \cite{biglan2000value, christian2010prenatal}).  
In SM, intervened units are paired with control units and those units without a sufficiently-good
match according to a given statistical criteria are excluded when estimating the treatment effect.
In the running example of animal-species classification, intervened units may correspond to the
limited set of camera trap locations that are fully-annotated, while control units refer to the
many more camera trap locations that are only partially annotated.
Pairing is beneficial for capturing diverse operating conditions, yet the ability to drop unpaired
units is crucial for mitigating the risk of statistically-poor matches corrupting the training
signal.

By developing an online method for statistically matching samples from different domains
(camera-trap locations) and using this to define a consistency loss, we arrive at our proposed
semi-supervised method, \emph{Okapi}. 
This consistency loss is predicated on the simple idea of pulling together similar samples from
different domains within the latent space of the encoder, and using this to bootstrap said encoder
such that the distributions become progressively more aligned over the course of training. 
Since matching samples using the full dataset at each step of training is computationally
infeasible, we instead approximate it using a combination of momentum-encoding and a memory-bank
that has been well-proven in self-supervised learning \citep{he2020momentum, koohpayegani2021mean}.
Compared with other consistency-based methods such as FixMatch \citep{sohn2020fixmatch}, Okapi has
the advantage of being agnostic to both the task and the modality, in addition to being
distributionally robust.
Contrary, to \citet{SagWeiLeeGaoetal22}, we show that the supplementary unlabelled data and domain
information can be leveraged by Okapi to improve upon standard ERM on datasets from the WILDS 2.0
benchmark.

%% file: preliminaries.tex
\subsection{Problem setting}

In the standard supervised setting, one is given a dataset, $\mc{D}_l \triangleq \{x_i,
y_i\}_{i=1}^{N_l}$, and trains a model, parameterised by $\theta$, to well-approximate the
empirical distribution as $p_\theta(y | x)$.
Labelled data is limited by the cost of annotation yet one often has access to a far larger corpus
of unlabelled data, $\D_u \triangleq \{x_i\}_{i=1}^{N_u}$, which can be used to supplement $\D_l$. 
Semi-supervised learning is motivated by the idea that this additional data can often be used to
improve the ID and/or OOD performance of $p_\theta(y | x)$.
We can view unsupervised domain adaptation (UDA) as a special case of semi-supervised learning,
where there is assumed to be some distribution shift (adverse to a na\"ively-trained predictor)
between $\D_l$ and $\D_u$. Here, $\D_u$ comes from the domain on which $p_\theta(y | x)$ is to be
evaluated, such that we have $\mc{D}_u \triangleq \D_\mr{OOD}$, where $\mc{D}_\mr{OOD}$ denotes the
target domain, that is OOD w.r.t. $D_l$.
In the most general sense, a \emph{domain}, or \emph{environment} \citep{arjovsky2019invariant,
creager2021environment} describes some partitioning of the data according to its source or some
secondary characteristic, such as time of day, weather, location, lighting, or the model of the
device used to collect said data; one would hope that a predictor trained under one set of
conditions (e.g. day) would perform with minimal degradation under another set of conditions (e.g.
night) when those conditions are irrelevant to the task at hand.

Assuming the data follows the conditional generative distribution $x \sim p(x | s)$, where $s$ is
the domain label, one would ideally use $\D_{\mr{OOD}}$ to learn invariance to the marginal
distribution, $p(s)$, and thereby achieve the equivalence $p_\theta(y | x) = p_\theta(y | x, s)$.
In practice, one typically does not have access to $\mc{D}_{\mr{OOD}}$ but does have access to
training data sourced from a mixture of domains which can be leveraged to learn a more general
invariance that extends to those domains outside the training distribution
\cite{arjovsky2019invariant}.
Such a learning paradigm is referred to as domain generalisation (DG).
While some DG works consider the more extreme case of $s$ being unobserved
\citep{creager2021environment}, we follow the more conventional setup \citep{arjovsky2019invariant,
krueger2021out, SagWeiLeeGaoetal22} in which the domain(s) associated with each sample (labelled and
unlabelled) is indicated by the discrete label (set of labels) $s$. 
We denote the set of possible domains for the in-distribution labelled and unlabeled data, as
$\mc{S}_l$ and $\mc{S}_u$, respectively, and their union as $\mc{S} \triangleq S_l \cup S_u$.
Following the setup established in \cite{SagWeiLeeGaoetal22}, $\D_u$ is assumed to be unlabelled
only w.r.t the targets and not w.r.t the domain labels and thus that both $\D_l$ and $\D_u$ can be
augmented with the latter to give the re-definitions $\D_l \triangleq \{ x_i, y_i, s_i
\}_{i=1}^{N_l}$ and $\D_u \triangleq \{x_i, s_i \}_{i=1}^{N_u}$.

\subsection{Statistical matching} \import{./}{matching.tex}

%% file: matching.tex
Statistical matching is a sampling strategy which aims to balance the distribution of the observed
covariates in the \emph{treated} and \emph{control} groups. In general terms, observed covariates
$x$ are measured characteristics of the samples; in our work we refer to the encodings generated by
a deep neural network as covariates instead of the original characteristics.
The treated and control groups are two partitions of the data; specifically, the treated group is
the set of samples having a specific value of a variable of interest (here, the domain indicator, $s$) and
the control group is its complement.

In this work we utilise Nearest Neighbour (NN) matching, a distance-based matching method that pairs
sample $i$ of the treated group with the closest sample $j$ belonging to the control group.
A distance measure is used to define how close two samples, $i$ and $j$, are, with \emph{propensity
score distance} (PSD) and \emph{Euclidean distance} being two widely-used distances that we rely on
-- indirectly (as a means of filtering) and directly, respectively -- in this work.

%

The propensity score distance is defined as the difference between propensity scores, $e_i$
and $e_j$, of samples $i$ and $j$, i.e. $\mr{PSD}(e_i, e_j) \triangleq \vert e_i - e_j \vert$. In
causal inference, the propensity score refers to the probability of sample $i$ belonging to the
treated group, given its covariates $x_i$ \cite{rosenbaum1983central}; in practice, this
conditional probability is rarely known a priori and thus requires estimation, typically via
logistic regression \cite{stuart2010matching}.
The Euclidean-distance approach, in contrast, computes the distance between the covariates,
$x_i$ and $x_j$, of a given pair of samples.
Despite PSD being the more prevalent of the two distances, it is ill-suited to cases
in where pairs are close in value w.r.t. all covariates and in such cases Euclidean distance should
be preferred
\cite{king2019propensity}. 
Nevertheless, propensity scores remain a relevant component of NN-based matching for defining
\emph{calipers} that can reduce the likelihood of false-positive matches.

In this work we make use of two types of caliper, \emph{fixed} and \emph{standard deviation}. The fixed
caliper \cite{crump2009dealing}, $t_f$, defines a region of common support between the estimated
propensity score distribution of the two groups; only those samples within the feasible region are
admissible for matching ($\left\{i | e_i \in (1 - t_{f}, t_{f}) \right\}$).
This selection rule helps by removing samples with extreme propensity scores.
The standard deviation-based caliper (std-caliper), on the other hand, \cite{rosenbaum1985constructing} determines the
maximum discrepancy that might exist between two samples while still being admissible for pairing.
The discrepancy is usually expressed in terms of estimated PSD: $|e_i - e_j| < \sigma \cdot t_\sigma$.
Here, $\sigma$ denotes the mean of the group-wise standard deviations of the propensity
scores and $t_\sigma$ is a parameter controlling the percentage of bias reduction of the covariates. 
In the following section, we describe how one can leverage this matching framework to define a
consistency loss encouraging inter-domain robustness.

%% file: methodology.tex
Here, we introduce \emph{Okapi}, a simple, efficient, and general (in the sense that it is
applicable to any task \emph{or} modality) method for robust semi-supervised learning based on
online statistical matching. 
Our method belongs to the broad family of consistency-based methods, characterised by methods such
as FixMatch \citep{sohn2020fixmatch}, where the idea is to enforce similarity between a model's
outputs for two views of an unlabelled image. 
These semi-supervised approaches based on minimising the discrepancy between two views of a given
sample are closely related with self-supervised methods based on instance discrimination
\citep{chen2020simple} and self-distillation \citep{baevski2022data2vec,caron2021emerging,grill2020bootstrap}.
Many of the methods within this family, however, are limited in applicability due to their
dependence on modality-specific transformations and only recently has research into
self-supervision sought to redress this problem with modality-agnostic alternatives such as MixUp
\citep{verma2021towards}, masking \citep{baevski2022data2vec}, and nearest-neighbours
\citep{dwibedi2021little, koohpayegani2021mean, van2021revisiting}.
Approaches such as FixMatch, AlphaMatch \citep{gong2021alphamatch} and CSSL
\citep{lienen2021credal} that enforce consistency between the \emph{predictive} distributions
suffer further from not being directly generalisable to tasks other than classification.
Okapi addresses both of these aforementioned issues through 1) the use of a statistical matching
procedure -- that we call \CNN\ and detail in Sec. \ref{subsec:matching} -- to generate
multiple views for a given sample; 2) enforcing consistency between encodings rather than between
predictive distributions.

We show that models trained to maximise the similarity between the encoding of a given sample and
those of its \CNN-generated match are significantly more robust to real-world distribution shifts
than the baseline methods, while  having the advantage of being both computationally efficient
and agnostic to the modality and task in question.
Qualitatively speaking, we see that matches produced with the final model are related in
semantically-meaningful ways. 
Furthermore, since the only constraint is that samples be from different domains, the method is
applicable whether information about the domain is coarse or fine-grained.

In the following subsections, we begin by giving a general formulation of our proposed
semi-supervised loss employing a generic cross-domain $k$-NN algorithm. 
We then explain how we can replace this algorithm with \CNN\ in order to mitigate the risk of
poorly-matched samples, and how the loss may be computed in an online fashion to give our complete
algorithm.

\subsection{Enforcing consistency between cross-domain pairs}

We consider our predictor as being composed of an encoder (or \emph{backbone}) network,
$f_\theta:
\mc{X} \rightarrow \mathbb{R}^d$, generating intermediary outputs (features) $z \triangleq
f_\theta(x)$, and a prediction head, $g_\phi$, such that the prediction for sample $x$ is given by
$\hat{y} \triangleq g_\phi \circ f_\theta(x)$.
We similarly consider the aggregate loss $\mc{L}$ as having a two-part decomposition given by
\begin{equation}
    \mc{L} \triangleq \mc{L}_{\mr{sup}} + \lambda \, \mc{L}_{\mr{unsup}},
\end{equation}
where $\mc{L}_\mr{sup}$ is the supervised component measuring the discrepancy (as computed, for
example, by the MSE loss) between $\hat{y}$ and the ground-truth label $y$, $\mc{L}_{\mr{unsup}}$
is the unsupervised component based on some kind of pretext task, such as cross-view consistency,
and $\lambda$ is a positive pre-factor determining the trade-off between the two components.
For our method, we do not assume any particular form for  $\mc{L}_{\mr{sup}}$ and focus solely on
$\mc{L}_\mr{unsup}$.

Given a pair of datasets $\mc{D}_l$ and  $\mc{D}_u$, sourced from the labelled domain $\mc{S}_l$,
and unlabelled domain $\mc{S}_u$ respectively, along with their union $\mc{D} \triangleq \mc{D}_l
\cup \mc{D}_u$ our goal is to train a predictor that is robust (invariant) to changes in domain,
including those unseen during training.
To do this, we propose to regularise $z \triangleq f_\theta(x)$ to be smooth (consistent) within
local, cross-domain neighbourhoods.
At a high-level, for any given \emph{query} sample $x_q$ sourced from domain $s_q$, we compute
$\mc{L}_{\mr{unsup}}$ as the mean distance between its encoding $z_q \triangleq f_\theta(x_q)$ and
that of its $k$-nearest neighbours, $V_k(z_q)$ with the constraint that $\{ s_q \} \cap \mbf{s}_n =
\emptyset$, where $\mbf{s}_n$ is the set of domain-labels associated with $V_k(z_q)$.
The general form of this loss for a given sample can then be written as
\begin{align}
    V_k(z_q) &\triangleq \mr{NN}(z_q, \{ f_\theta(x)\;|\; (x, s) \in \mc{D}, s \neq s_q \}, k),
    \label{cgnn} \\
    \mc{L}_{\mr{unsup}} &\triangleq \frac{1}{k} \sum_{z_n \in V_k(z_q)} d(z_q, z_n)
\end{align}
where $d: \mathbb{R}^d \times \mathbb{R}^d \rightarrow \mathbb{R}$ is some distance function.
Here, we follow \cite{grill2020bootstrap} and define $d$ to be the squared Euclidean distance
between normalised encodings for our experiments.
Allowing the $\mr{NN}$ algorithm to select pairs in an unconstrained manner, given the pool of
queries and keys, however, can lead to poorly-matched pairs that are detrimental to the
optimisation process.
To address this, we replace the standard $\mr{NN}$ algorithm with a propensity-score-based variant,
inspired by the statistical matching framework \citep{rosenbaum1983central}.

\subsection{Cross-domain matching}\label{subsec:matching}

\begin{figure}[tbp]
  \centering
  \includegraphics[width=1.\textwidth]{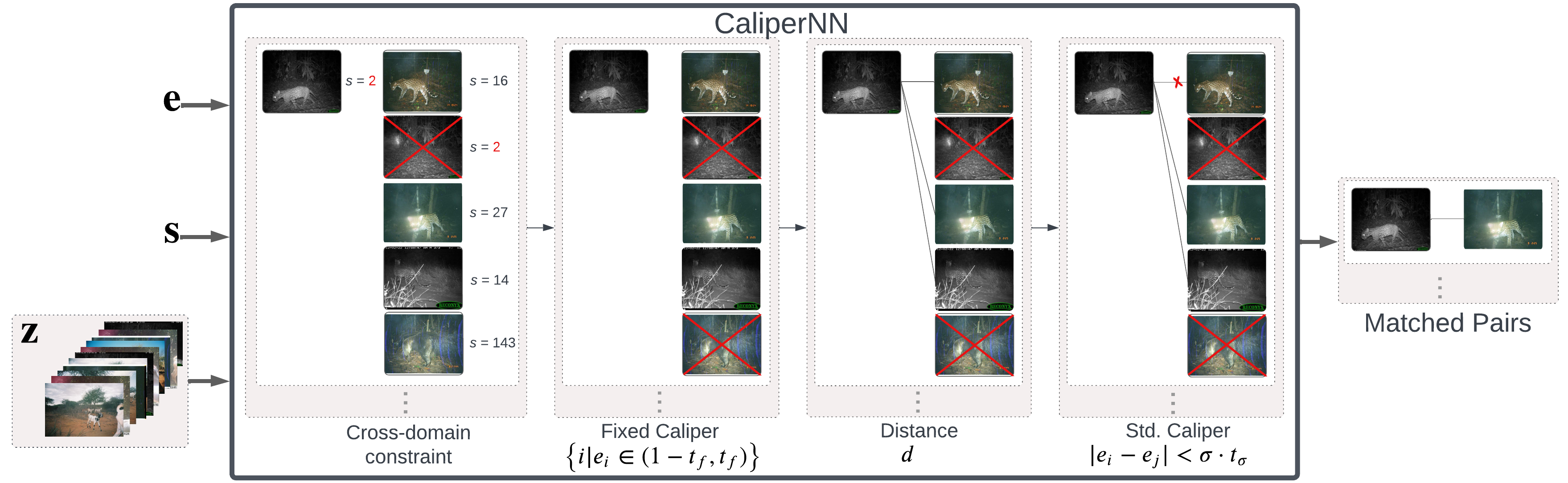}
  \caption{
 Illustration of our proposed statistical matching algorithm, \CNN. Given the anchor image
 encoding $\mbf{z}$, the corresponding domain label $s$, and propensity score $e$, \CNN~outputs the
 closest samples, according to distance $d$, subject to their being from different domains to the
 anchor.
}
  \label{fig:calipernn_pipeline}
\end{figure}

For the matching component of our algorithm, we propose to use a variant of $k$-NN which, in
addition to incorporating the above cross-domain constraint, filters the queries and keys that
represent probable outliers, according to their learned propensity scores.
The initial stage of filtering employs a fixed caliper, where samples with propensity scores
surpassing a fixed confidence threshold are removed; this is followed by a second stage of
filtering wherein any two samples (from different domains) can only be matched if the Euclidean
distance between their respective propensity scores is below a pre-defined threshold (std-caliper).
See Fig~\ref{fig:calipernn_pipeline} for a pictorial representation of these steps and
Appendix~\ref{appx:pseudocode} for reference pseudocode.

The propensity score, $e$, for a given sample $x$ is estimated as $p(s|z)$ using a linear
classifier $f_\theta$, $h_\psi: \mathbb{R}^d \rightarrow \bigtriangleup^{|\mc{S}|}$ where
$\bigtriangleup^{|\mc{S}|}$ is the probability simplex over possible domain labels, $\mc{S}$.
$h_\psi^d$ is trained via maximum (weighted) likelihood to predict the domain label of a given
sample for all samples within the aggregate dataset $\mc{D}$, or (typically) a subset of it,
encoded by $f_\theta$.
Since we apply both calipers to the learned propensity score, the shape of this distribution can
have a significant effect on the outcome of matching.
Accordingly, we apply temperature-scaling, with scalar $\tau \in \mathbb{R}^+_\star$, to sharpen or
flatten the learned propensity-score distribution such that we have $e \triangleq \mr{softmax}(
\tau^{-1} \, h_\psi(z))$.
We denote the set of associated parameters (\{ $t_{f}, t_{\sigma}, \tau \}$, as the threshold for
the fixed-caliper, the threshold for the std-caliper, and the temperature, respectively) as $\xi$
and discuss in Appendix~\ref{appx:implementation} how one can determine suitable values for these in practice.

For convenience we define the set of all encodings, given by $f_\theta$, as $\mbf{z} \triangleq \{
f_\theta (x)\, |\, x \in \mc{D} \}$, the set of all associated propensity scores as $\mbf{e}
\triangleq \{ h_\psi (x) \, | \, z \in \mbf{z} \}$, and the set of associated domain labels as
$\mbf{s}$.
In the offline case, the matches for $\D$ are then computed as

\begin{equation}
    \mr{MatchedSamples} \triangleq \{ (z,  \mr{\CNN}_\xi(z, \mbf{z}, \mbf{e}, s, k)) \, | \, z \in
    \mbf{z}, s \in \mbf{s} \},
\end{equation}

with \CNN \ returning the set of $k$-nearest neighbours according to $d$, subject to the
aforementioned cross-domain and caliper-based constraints. We allow for the fact that there may be
no valid matches for some samples due to these constraints; in such cases we have $\emptyset$ as
the second element of their tuples, indicating that $\mc{L}_{\mr{unsup}}$ should be set to $0$.

\subsection{Scaling up with Online Learning}\label{subsec:ol}

Re-encoding the dataset following each update of the feature-extractor, in order to recompute
$\mr{MatchedSamples}$, is prohibitively expensive, with cost scaling linearly with $N \triangleq
N_l + N_u$.
Moreover, \CNN \ requires explicit computation of the pairwise distance matrices, which can be
prohibitive memory-wise for large values of $N$.
We address these problems using a fixed-size memory bank, $\M^{\NM}_z$ storing only the last $\NM$
(where $\NM \ll N$) encodings from a slow-moving momentum encoder \citep{grill2020bootstrap,
he2020momentum}, $\textcolor{tecol}{f_{\theta^\prime}}$, which we refer to as the
\textcolor{tecol}{\emph{target}} encoder, in line with \cite{grill2020bootstrap}, and accordingly
refer to $\textcolor{oecol}{f_\theta}$ as the \textcolor{oecol}{\emph{online}} encoder.
Unlike \cite{grill2020bootstrap}, however, we make use of neither a projector nor a predictor head
(in the case of the target encoder) in order to compute the inputs to the consistency loss and
simply use the output of the backbone as is -- this is possible in our setting due to
$\mc{L}_\mr{sup}$ preventing representational collapse.
More specifically, the target encoder's parameters, $\textcolor{tecol}{\theta^\prime}$, are computed as a moving
average of the online encoder's, $\textcolor{oecol}{\theta}$, with decay rate $\zeta \in (0, 1)$, per the recurrence
relation
\begin{equation}
  \textcolor{tecol}{\theta^\prime_t} = \zeta \textcolor{tecol}{\theta^\prime_{t - 1}} + (1 - \zeta)
  \textcolor{oecol}{\theta_t},
\end{equation}
As the associated domain labels are also needed both for matching and to compute the loss for the
propensity scorer, we also store the labels associated with $\M^{\NM}_z$ in a companion memory bank
$\M^{\NM}_s$. 
We initialise $\M^{\NM}_z$ and $\M^{\NM}_s$ to $\emptyset$, resulting in
fewer than $\NM$ samples being used during the initial stages of training when the memory banks are
yet to be populated.

Each iteration of training, we sample a batch of size $B$ from $\D$ consisting of inputs $\mbf{x}$
and $\mbf{s}$.
During the matching phase, the inputs are passed through the \emph{target} encoder to obtain
$\mathbf{z}_q^\prime \triangleq \{ \textcolor{tecol}{f_{\theta^\prime} } (x) | x \in \mbf{x}\}$,
serving as the queries for \CNN.
We also experiment with a simpler variant where the \emph{online} encoder is instead used for this
query-generation step, such that we instead have $\mathbf{z}_q^\prime \triangleq
\{\textcolor{oecol}{f_{\theta}}(x) | x \in \mbf{x}\}$, and find this can work equally well if
$\zeta$ is sufficiently high.
The keys are then formed by combining the current queries with the past queries contained in the
memory bank: $\mbf{z}_k\triangleq \mathbf{z}_q^\prime \cup \M^{\NM}_z$.
The domain labels associated with $\mbf{z}_k$ are likewise formed by concatenating the domain
labels in the current batch with those stored in $\M^{\NM}_s$: $\mbf{s}_k \triangleq \mbf{s}_q \cup
\M^{\NM}_s$.
Once the matches for the current samples have been computed, the oldest $B$ samples in $M^{\NM}_z$
and $M^{\NM}_s$ are overwritten with $\mathbf{z}_k$ and $\mathbf{s}_k$, respectively.
The consistency loss is then enforced between each query $\mathbf{z}_q \triangleq \{
  \textcolor{oecol}{f_{\theta}}(x) |
x \in \mbf{x}\} $, according to the differentiable \emph{online} encoder, and each of its matches,
$V_k(z_q^\prime) \triangleq \mr{\CNN}_\xi(z_q, \mbf{z}_k, h_\psi(\mbf{z}_k), \mbf{s}_k)$ providing
that $V_k(z_q^\prime) \neq \emptyset$ (that is, under the condition that the estimated propensity
score for $z_q^\prime$ does not violate the caliper(s) and there are at least $k$ valid matches
whose estimated propensity scores also do not), with the loss simply $0$ otherwise.
Since $\textcolor{tecol}{ f_{\theta^\prime} }$ is frozen, $\mbf{z}_k$ carries an implicit stop-gradient and gradients
are computed only w.r.t. $\textcolor{oecol}{ \theta}$.
These steps are illustrated pictorially in Fig~\ref{fig:pipeline} and as pseudocode in
Appendix~\ref{appx:pseudocode}.

Similarly, rather than solving for the optimal parameters, $\psi^\star$ for the propensity scorer
given the current values of $\mbf{z}_k$, which is infeasible for the large values of $\NM$ needed
to well-approximate the full dataset, we resort to a biased estimate of $\psi^\star$.
Namely, we train $h_\psi$ in an online fashion to minimise the per-batch loss 
\begin{equation}
\mc{L}_\mr{ps} = \frac{1}{|\mbf{z}_k|} \sum_{ z \in \mbf{z}_k, s \in \mbf{s}_k } w_{\mbf{s}_k}(s) \mc{H}(h_{\psi}(z), s), 
\end{equation}
where $\mc{H}$ is the standard cross-entropy loss between the predictive distribution and the
(degenerate) ground-truth distribution, given by the one-hot encoded domain labels, and
$w_{\mbf{s}_k}: \mc{S} \rightarrow \mathbb{R}^+_*$ is a function assigning to each $s$ an importance weight
\citep{shimodaira2000improving} based on the inverse of its frequency in $\mbf{s}_k$ to counteract label imbalance.
In the special case in which the $\D_l$ and $\D_u$ are known to have disjoint support over $S$
(that is, $\mc{S}_l \cap \mc{S}_u = \emptyset$), we can substitute their domain labels with $1$ and
$0$, respectively (such that we have $\D_l \triangleq \{ x_i, y_i, 1 \}_{i=1}^{N_l}$ and $\D_u
\triangleq \{x_i, 0 \}_{i=1}^{N_u}$), thus reducing the propensity scorer and \CNN\ to their binary
forms. 
Knowing whether this condition is satisfied a priori (and thus whether the use
of domain labels can be forgone completely from our pipeline) is not unrealistic: one may, for
example know that two sets of satellite imagery cover two different parts of the world (e.g. Africa
and Asia) yet not know the exact coordinates underlying their respective coverage.

\begin{figure}[ht]
  \centering
  \includegraphics[width=1\textwidth]{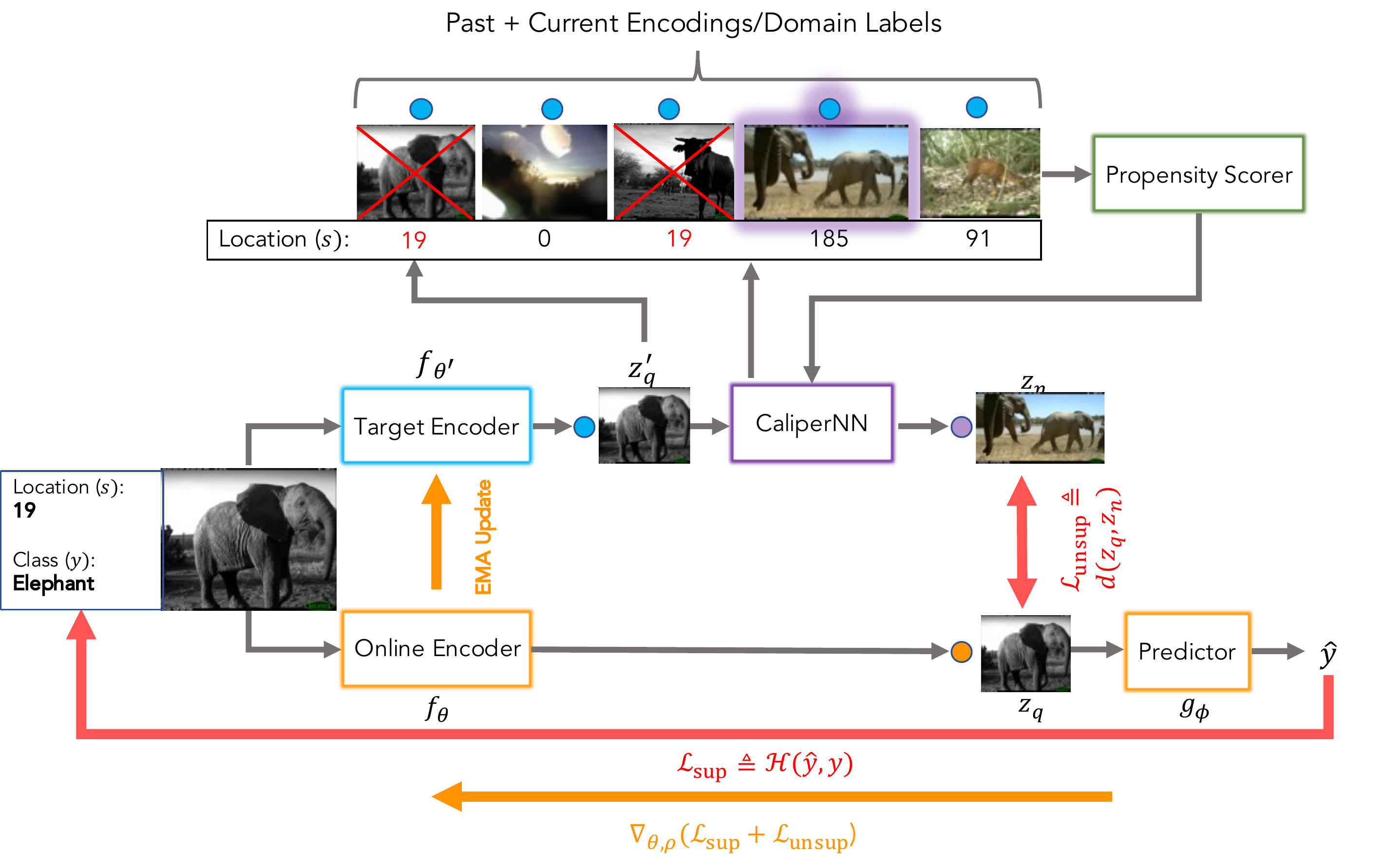}
  \caption{
  Overview of Okapi's online-learning pipeline based using the iWildCam dataset for the sake of
  illustration.
  For simplicity, we limit $k$ to 1 so that the output of matching is a single vector rather than a
  set of vectors; for the same reason we illustrate the process for only a single sample 
  taken from the labelled data set $\mc{D}_l$, annotated with both domain ($s$; in this case, \emph{camera
  location}) and class ($y$) information.
  Inspired by recent advances in self-supervised learning, we maintain a copy (the target encoder)
  of the online encoder, $f_\theta$, whose parameters, $\theta^\prime$, are an exponential moving
  average (EMA) of $\theta$. 
  This EMA update is performed at the beginning of each training set at a rate governed by the
  decay coefficient, $\zeta$. 
  For a given sample, we first compute its embedding using the target encoder to
  produce the query vector, $z_q^\prime$, and by the online encoder to produce $z_q$, which will
  serve as the 'anchor' in the consistency loss. 
  This query vector is then used -- alongside the output of the propensity scorer -- by \CNN\ to
  compute its cross-domain nearest neighbour, $z_n$, where the keys are taken to be the current and
  past (stored in the Memory Bank) $N_\mc{M}$ encodings of the data.
  The cross-domain constraint, prohibiting matching of samples belonging to the same domain,
  is denoted through a red coloring of the location identifiers, the nearest sample obeying
  this constraint and the constraints of the calipers with purple highlighting.
  The consistency loss is the distance between $z_q$ and $z_n$, defined by function some distance
  function $d$. 
  Finally, the supervised loss, $\mc{L}_\mr{sup}$ (here instantiated as the standard cross-entropy loss, $\mc{H}$), is computed using the output of the predictor
  acting on $z_q$ and the ground-truth given by $y$.
  }
  \label{fig:pipeline}
\end{figure}

%% file: related_work.tex
\paragraph{Domain Generalisation} 
The goal of domain generalisation (DG) is to produce models that are robust to a wide range of
distribution shifts (including those outside the training distribution), given a training set
consisting of samples sourced from multiple domains.
Despite the various techniques (many well theoretically-motivated) designed to improve the
generalisation of deep neural networks current methods continue to fall short in the face of
natural distribution shifts \citep{gulrajani2020search, koh2021wilds}.
Indeed, ERM has repeatedly shown to be a strong baseline -- frequently outperforming dedicated
methods that leverage domain information or additional unlabelled data -- for DG
\citep{gulrajani2020search, SagWeiLeeGaoetal22}, despite the theoretical problems associated with
using it when the training and test sets are misaligned.
Until now, only pre-training on larger, more diverse datasets (with harder examples), has
consistently proven to improve OOD generalisation, yet allowing pre-trained models to fit the ID
data too closely can undo any such benefit conferred by the pre-training
\citep{andreassen2021evolution, kim2022broad, taori2020measuring, wiles2022a}.
%
%
Similar to Okapi, MatchDG \citep{mahajan2021domain} draws upon causal matching to tackle DG.
Despite the surface-level similarity, there are a number of significant differences, principally in the
respects that we consider semi-supervised DG (whereas MatchDG requires full-labeling w.r.t.
$y$) and employ an augmented form of k-NN for bias-reduction in the absence of $y$.

\paragraph{Self-Supervised Learning} 
In self-supervised learning (SelfSL), models are trained to solve pretext tasks constructed from
the input data.
This learning paradigm has led to significant breakthroughs in unsupervised learning in recent
years, with performance now approaching (or even surpassing, along some axes such as adversarial
robustness) that of supervised methods for many tasks while requiring significantly less labelled data.
Due to its generality, SelfSL has seen use across the complete spectrum of applications and
modalities and underlies many of the foundation models \citep{bommasani2021opportunities} that have
emerged in NLP \citep{brown2020language, chowdhery2022palm, devlin2018bert}, Computer Vision
\citep{goyal2022vision}, and at their intersection \citep{alayrac2022flamingo, yu2022coca}.
Common pretext tasks include those based on the masked-language-modelling approach -- originally
popularised by BERT \citep{devlin2018bert} and recently generalised to other modalities
\citep{baevski2022data2vec, bao2021beit} -- \citep{chen2020simple, he2020momentum}, contrastive
captioning  \citep{radford2021learning, yu2022coca}, and instance discrimination and
self-distillation \citep{caron2021emerging, grill2020bootstrap} which rely on transformations of
the data to generate multi-view inputs.
Approaches belonging to the latter two categories were originally limited by the fact that the
transforms had to be tailored for a particular modality and for some modalities, such as tabular
data, there is no obvious way to define them.
A number of recent works have sought to obviate this problem through the use of  MixUp
\citep{verma2021towards}, masking \citep{baevski2022data2vec, MaskedAutoencoders2021}, and k-NN
\citep{dwibedi2021little, koohpayegani2021mean, van2021revisiting}, the latter of which is directly
relevant to our work.
Okapi bears closest resemblance to \cite{koohpayegani2021mean} in combining momentum-encoding with
nearest-neighbours lookup to generate the views for a BYOL-style \citep{grill2020bootstrap}
consistency loss. 
However, a key distinction lies in the use of an augmented form of nearest-neighbours, \CNN, which
both constrains pairs of samples to being from \emph{different} domains and filters out any queries
or keys deemed outliers according to a learned \emph{propensity score}. 

\paragraph{Semi-Supervised Learning}  
Semi-supervised learning (SemiSL) encompasses a broad class of algorithms that combine unsupervised
learning with supervised learning in order to improve the performance of the latter, especially
when labelled data is limited.
Many SemiSL methods are based on the self-training paradigm which can trace its roots back decades
to the early work in pattern recognition by \cite{scudder1965probability} and continues to be
relevant in the modern era due to its generality, both within SemiSL itself and in related fields
such as domain adaptation \citep{ganin2016domain}, and fledgling field of SelfSL
\citep{caron2021emerging} discussed above.
Self-training applies to any framework predicated on using a model's own predictions to produce
pseudo-labels for the unlabelled data which can either be used as targets for self-distillation
\citep{xie2020self} or enforcing consistency between predictions that themselves have been
perturbed \citep{bachman2014learning, xie2020self} or that have been generated from
perturbed/multi-view inputs \citep{sohn2020fixmatch}.
FixMatch \citep{sohn2020fixmatch} is one example of a consistency-based method which has proven
effective for semi-supervised classification, despite its simplicity, and various works
\citep{gong2021alphamatch, lienen2021credal} have since built on the its framework prescribing the
use of weakly- and strongly-augmented inputs to generate the targets and predictions, respectively.
Like these methods, Okapi also makes use of a cross-view consistency loss, however, the alternative
views for a given sample are generated not through data-augmentation but through statistical
matching \citep{rosenbaum1983central}, with the aim being to achieve invariance to the domain
rather than a particular series of perturbations.
Another example of particular relevance to our work is \cite{tarvainen2017mean}, which uses a copy
of the model with exponentially-averaged weights to generate the targets for the unlabelled data.
Okapi also uses such a model to produce the targets for its consistency loss, but is more akin to
momentum-encoding \citep{he2020momentum} in the respect that the loss is imposed on the latent
space.

%% file: experiments.tex
\subsection{Datasets}\label{sec:exps_datasets} We evaluate Okapi on three datasets taken from the
WILDS 2.0 benchmark \cite{SagWeiLeeGaoetal22}. These span a variety of modalities and tasks,
allowing us to showcase the generality of our proposed method (Okapi): \textbf{iWildCam} (images,
multiclass classification), \textbf{PovertyMap} (multispectral images, regression), and
\textbf{CivilComments} (text, binary classification). 
Details of each dataset can be found in Appendix~\ref{appx:datasets}.

\subsection{Image experiments}

\import{./}{tables/iw_pm_results.tex}

Results of our image-data experiments are summarised in Table~\ref{tab:iw_pm_results}. Due to spacial 
constraints, we defer the full set of results, including those for the `offline' (w.r.t. the
matching) version of Okapi to Appendix.~\ref{appx:ext_results}.
For both datasets in question, we use the same metrics as \cite{SagWeiLeeGaoetal22}: macro-F1 for
iWildCam and worst-group (with the group defined as urban (U) vs. rural (R)) Pearson correlation
for Poverty Map. For completeness, we include mean squared error (MSE) as a secondary metric for
the latter dataset. Following \cite{SagWeiLeeGaoetal22}, we compute the mean and standard deviation
(shown in parentheses) over multiple runs for both ID and OOD test sets, with these runs conducted
with 3 different random seeds and 5 pre-defined cross-validation folds for iWildCam and
PovertyMap, respectively.

We compare Okapi against two baselines, ERM and FixMatch \citep{sohn2020fixmatch},
both according to our re-implementation and according to the original implementation given in
\cite{SagWeiLeeGaoetal22}.
We note that since FixMatch, in its original form, is only applicable to classification problems
due to its use of confidence-based thresholding, for the PovertyMap dataset, FixMatch represents a
simplified variant (following \citep{SagWeiLeeGaoetal22}) without such thresholding, that is
trained to simply minimise the MSE between \emph{all} regressed values for the weakly- and
strongly-augmented images.
As described in Appendix~\ref{appx:implementation}, the main difference between the baselines run included in
\cite{SagWeiLeeGaoetal22} and our re-runs is in the backbone architecture, with us opting for a
ConvNeXt  \citep{liu2022convnet} architecture over a ResNet one.
For both datasets, and for both baselines we observe significant improvements stemming the change
of backbone.
%
%
Moreover, utilising ConvNeXt seems to be crucial in enabling FixMatch to surpass the ERM baseline
in the classification task with $32.2$ (ERM) vs $31.0$ (FixMatch) and $33.3$ (ERM) vs. $35.2$
(FixMatch), with ResNet and ConvNeXt architecture respectively. 

Okapi, convincingly outperforms the baselines, w.r.t the OOD metric of interest, on both datasets. 
We observe an improvement of $+0.9$ macro F1, i.e. $36.1$ vs $35.2$ of Okapi and FixMatch (the best
baseline for iWildCam) respectively. For the regression task in PovertyMap, Okapi achieves $0.55$
and $0.33$ on the OOD test set in terms of Pearson correlation and MSE, respectively, in contrast
to the $0.53$ and $0.33$ of ERM.
At the same time, we note that FixMatch fails to generalise well to this task, yielding by far the
worst results amongst the evaluated methods.

\subsection{Text classification}

\import{./}{tables/cc_results.tex}

In Table~\ref{tab:cc_results} we summarise the numerical results for the CivilComments dataset.
Remaining consistent with \cite{SagWeiLeeGaoetal22}, we evaluate models according to the
worst-group accuracy -- the minimum of the conditional accuracies obtained by conditioning on each
of the 8 dimensions of $s$ -- averaged over 5 replicates. Since there is no canonical ID test split
available for this dataset, we report only the results only for the OOD split that is, rather than
doing so for a custom split to avoid misrepresentation. We compare Okapi against both ERM variants
featured in \cite{SagWeiLeeGaoetal22} -- one trained on only the official labelled data and one
trained with annotated unlabelled data (fully-labelled) -- as well as our re-implementation of the
ERM variant trained on only the labelled data with an identical hyperparameter configuration to the
former. In contrast to the image datasets, we do not diverge in our choice of architecture, with
all models trained with a pre-trained DistilBERT \citep{sanh2019distilbert} backbone.

We observe marked improvement in the worst-group accuracy of this baseline compared with that
reported therein. We attribute this partly to the high variance of the model-selection procedure
(inherited from \cite{SagWeiLeeGaoetal22}) based on intermittently-computed validation performance
(which does not consistently align with test performance) to determine the final model. This aside,
we observe that Okapi outperforms the ERM baseline by a significant margin, to the point of parity
with the fully-labelled baseline.

\subsection{Ablations and qualitatitive analysis}

In order to evaluate the importance of the caliper-based filtering to the performance of Okapi, we
perform an ablation experiment on PovertyMap dataset (Okapi (no calipers)) with said filtering
disabled (and all else constant), such that instead of \CNN\ we have standard $k$-NN, albeit with
the cross-group constraint still in place (per Eq.~\ref{cgnn}).
We see that performance degrades according to both metrics of interest, and, crucially, that the
standard deviation of the runs is significantly higher, in line with our expectation that filtering
out poor matches should stabilise optimisation.
We provide additional ablation experiments in Appendix~\ref{appx:ablations}, exploring the relative
importance of the two (fixed and std-) calipers, the optimal number of neighbours to use for
computing $\mc{L}_\mr{unsup}$, and the feasibility of using the online encoder to generate the
queries for \CNN.

\begin{figure}[tbp]
  \centering
  \includegraphics[width=1.\textwidth]{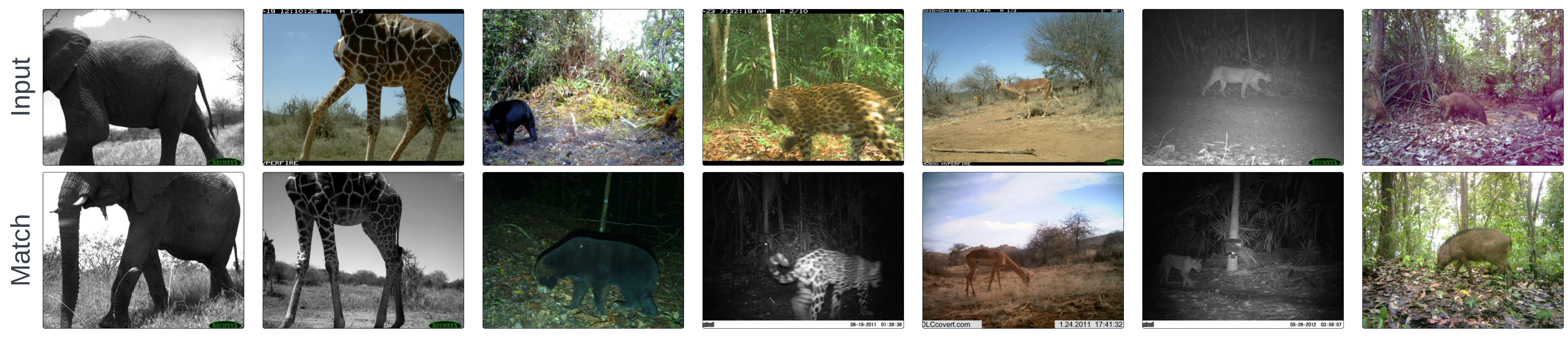}
  \caption{Examples of input (labelled) images and their 1-NN matched (unlabelled) images retrieved using \CNN\ on iWildCam dataset. Here, we match images from the labelled-train set to images from the unlabelled-extra set, taking advantage the fact that their domains are disjoint.
  }
  \label{fig:matches_examples}
\end{figure}

Finally, in Fig. \ref{fig:matches_examples} we show samples of matched pairs retrieved by \CNN\
from the encodings of the learned encoder for the iWildCam dataset.
Here, we see that semantic information (encoding the species of animal) is preserved across pairs,
while nuisance factors such as illumination, background and contrast vary.
Further examples from PovertyMap are shown in Appendix~\ref{appx:additional_matches}. In
Appendix~\ref{appx:pacs_matching}, we include matching results for the PACS (photo (P), art painting (A), cartoon (C), and
sketch (S))
dataset~\cite{li2017deeper} demonstrating how temperature scaling, in conjunction with the fixed
caliper, can be used to control the filtering rate.

%% file: tables/iw_pm_results.tex
\begin{table}[tbp]
	\centering
	\caption{
		A comparison between Okapi and different baselines on two benchmark image datasets.
		We include both the results of our re-run of the baselines and those of
		\cite{SagWeiLeeGaoetal22}.
		Both ID and OOD performances are reported. For iWildCam we average over results
		from 3 different seeds, for PovertyMap we do so over the 5 pre-defined CV folds.
		Standard deviations are shown in parentheses.
}
	\scalebox{0.90}{
	\begin{tabular}{lllllll}
		\toprule \textbf{Method} & \multicolumn{2}{c}{\textbf{iWildCam}} &
		\multicolumn{4}{c}{\textbf{PovertyMap}} \\ & \multicolumn{2}{c}{macro F1
		$\uparrow$} & \multicolumn{2}{c}{worst U/R corr. $\uparrow$} &
		\multicolumn{2}{c}{worst U/R MSE $\downarrow$} \\
	\midrule
        & ID & OOD & ID & OOD & ID & OOD \\
        ERM \cite{SagWeiLeeGaoetal22} & 47.0 (1.4) & 32.2 (1.2) & 0.66 (0.04) & 0.49 (0.06) & - & - \\
        FixMatch \cite{SagWeiLeeGaoetal22} & 46.3 (0.5) & 31.0 (1.3) & 0.54 (0.10) & 0.30 (0.11) & - & - \\
	\midrule
        ERM & 48.6 (1.1) & 33.3 (0.3) & 0.72 (0.03) & 0.53 (0.09) &  0.23 (0.03) & 0.35 (0.12)\\
        FixMatch & 51.1 (1.0) & 35.2 (0.7) & 0.50 (0.13) & 0.34 (0.12) & 0.59 (0.42) & 0.88 (0.61)\\
        \rowcolor{lightgray}
        Okapi (ours) & 50.6 (0.7) & 36.1 (0.9) & 0.72 (0.02) & 0.55 (0.10) & 0.22 (0.02) & 0.33 (0.10)\\
        Okapi (no calipers) & - & - & 0.72 (0.02) & 0.54 (0.12) & 0.22 (0.02) & 0.36 (0.14) \\
		\bottomrule
	\end{tabular}}
	\label{tab:iw_pm_results}
\end{table}

%% file: tables/cc_results.tex
\begin{SCtable}[][ht]
	\centering
	\caption{
	  Comparison between Okapi and the baselines methods on the Civil Comments dataset. We
	  include both the original results of \cite{SagWeiLeeGaoetal22} as well as those of
	  our reproduction of their ERM baseline. 
	  Performance is measured in terms of worst-group accuracy and averaged over seeds;
	  standard deviations are shown in parentheses.
	}
	\scalebox{0.90}{
	\begin{tabular}{llll} \toprule \textbf{Method} & \textbf{Civil Comments} \\ & worst-group
	  acc $\uparrow$
	\\ \midrule
	& OOD \\
        ERM \cite{SagWeiLeeGaoetal22} & 66.6 (1.6) \\
        ERM (fully-labelled) \cite{SagWeiLeeGaoetal22} & 69.4 (0.6) \\
	ERM (reproduction) & 68.5 (2.2) \\
        \rowcolor{lightgray}
	Okapi (ours) & 69.7 (2.0) \\
	\bottomrule
	\end{tabular}
      }
	\label{tab:cc_results}
\end{SCtable}

%% file: conclusion.tex
In this work, we introduced, Okapi, a semi-supervised method for training distributionally-robust
models that is intuitive, effective, and is applicable to any modality or task.
Okapi is based on the simple idea of supplementing the supervised loss with a cross-domain
consistency loss that encourages the outputs of an encoder network to be similar for neighbouring
(within the latent space of the encoder itself) samples  belonging to different domains, which is
made efficient using an online-learning framework.
Rather than simply using $k$-NN with a cross-domain constraint, however, we propose an augmented
form based on statistical matching (\CNN) that combines propensity scores with calipers to winnow
out low-quality matches; we find this to be important for both the end-performance and consistency
of Okapi.
Our work serves as a response to \cite{SagWeiLeeGaoetal22}, in that we find that it is in fact
possible to effectively incorporate unlabelled data and domain information into a training
algorithm in order to improve upon ERM with respect to an OOD test set, assuming an appropriate
choice of architecture.
Namely, on three datasets from the WILDS 2.0 benchmark, representing two different tasks
(classification and regression) and modalities (image and text), we show that Okapi outperforms
both the ERM and FixMatch baselines according to the relevant OOD metrics.

Buoyed by these promising results, we intend to apply Okapi to other tasks (e.g. object detection
and image segmentation) and other modalities (e.g. audio) to further establish its generality.
Furthermore, one limitation of the current incarnation of the method is that the thresholds for the
calipers are fixed over the course of training whereas it may be beneficial to set these adaptively
with the view to optimise such measures of inter-domain balance as \emph{Variance Ratio} and
\emph{Standard Mean Differences} that are commonly used to evaluate the the goodness of statistical
matching procedures.

\begin{ack} This research was supported by a European Research Council (ERC) Starting Grant for the
  project "Bayesian Models and Algorithms for Fairness and Transparency", funded under the European
  Union's Horizon 2020 Framework Programme (grant agreement no. 851538). NQ is also supported by
  the Basque Government through the BERC 2018-2021 program and by Spanish Ministry of Sciences,
  Innovation and Universities: BCAM Severo Ochoa accreditation SEV-2017-0718. \end{ack}

%% file: supplementary.tex
\appendix
\section{Datasets}\label{appx:datasets}
\import{./}{datasets.tex}

\section{Relation to Algorithmic Fairness}\label{appx:fairness}
\import{./}{fairness.tex}

\section{Extended Results}\label{appx:ext_results}

We tabulate in Table~\ref{tab:ext_iw_pm_results} an extended version of the results presented in
the main text. This includes additional results with the ResNet backbones per
\cite{SagWeiLeeGaoetal22} (justifying our decision to adopt a ConvNext backbone for our main set of
image-dataset results) as well as those for an 'offline' version of Okapi (Okapi (offline)) where
the matches are generated prior to training using features of the respective ERM baseline for each
dataset. Since the target encoder is necessitated by the need for online match-retrieval, only a
single encoder is involved in  Okapi (offline); in binary cases, the algorithm is then identical to
the one proposed by \cite{RomInsShaQua22} with the exception that consistency is still enforced via
distance in encoding space rather than with a JSD loss on the predictive distributions which fails
to generalise to regression tasks such as PovertyMap.

\import{./}{ext_results.tex}

\section{Implementation details}\label{appx:implementation}
\import{./}{implementation.tex}

\section{Additional Matching Examples}\label{appx:additional_matches}

\begin{figure}[ht!]
  \centering
  \includegraphics[width=1.\textwidth]{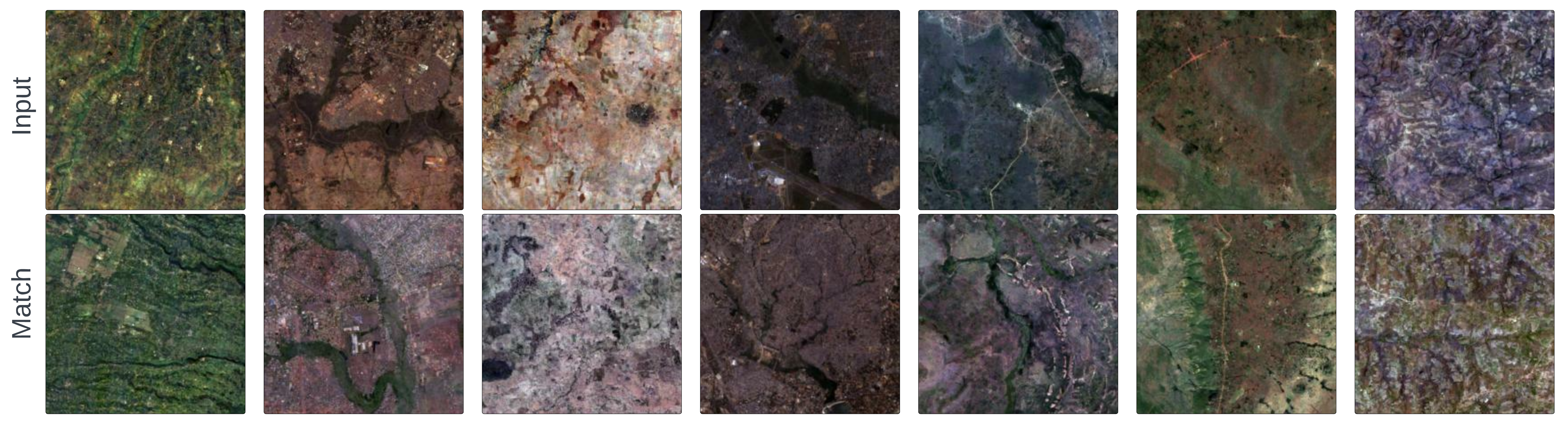}
  \caption{Examples of input (labelled) images and their 1-NN matched (unlabelled) images retrieved using \CNN\ from the PovertyMap-WILDS dataset. Here, we match images from the labelled-train set to images from the OOD-validation set, taking advantage the fact that their domains are disjoint.
  }
 \end{figure}
 
\section{Ablations}\label{appx:ablations}

\import{./}{tables/pm_ablations.tex}

 We supplement the ablation experiment on the use of calipers featured in the main text with
 additional experiments concerning effect of the number of nearest neighbours ($k$), the relative
 importance of the two (fixed and std) calipers, and the feasibility of using the online encoder as
 the query-generator instead of the target encoder.
 The results of these experiments are tabulated in \ref{tab:ablations} with the key takeaways
 being: \begin{enumerate} 
     \item The number of neighbours used for computing the consistency loss
         has little impact -- according to the given level of precision -- on the performance of Okapi
        along all axes.
     \item While disabling the calipers altogether considerably harmed performance, using only the
         std. caliper allows us to recover the performance of the complete algorithm,
         (\texttt{Okapi (k=5)}), whereas the same is not true for the fixed caliper which, while
         aiding performance compared to the no-caliper baseline, falls short of that benchmark.
         A caveat attached to these conclusions, however, is that the selected values for $\xi$ are
         likely suboptimal in the online setting, given that they were optimised for the static
         setting: with improved selection of $\xi$, either by learning it jointly with the model's
         parameters (using, for instance, the perturbed maximum method \citep{berthet2020learning}
         to overcome the non-differentiability of the $k$-NN and thresholding operations), in an
         amortised
     fashion, or optimising it on a per-iteration basis.
    \item While less appealing from a conceptual standpoint, due to the mismatch between the
        networks used to generate the queries and keys, from an empirical standpoint it is
        perfectly feasible to use the online encoder to generate the queries for statistical
        matching instead the target encoder while experiencing minimal degradation in performance.
     This is particularly relevant when one wishes to perform the matching in only one direction
(e.g. $\D_l \rightarrow \D_l$) due to the reduction in redundant encoding, with each encoder only
encoding its respective subset of the data (e.g. $f_\theta$ only encodes samples from $\D_l$ and
$f_\theta^\prime$ only encodes samples from $\D_u$) \end{enumerate}

\section{Pseudocode}\label{appx:pseudocode}

\import{./pseudocode/}{calipernn.tex}
\import{./pseudocode/}{ol.tex}

We provide Pytorch-style \citep{paszke2019pytorch} pseudocode for the \CNN\
(described in~\ref{subsec:matching}) and online-learning (described in~\ref{subsec:ol}) algorithms
in Algorithm~\ref{alg:calipernn_pc} and Algorithm~\ref{alg:ol_pc}, respectively.
In both cases, we restrict the pseudocode to the special case of binary domains --
 practically achieved by using the labelled/unlabelled as a proxy for domain -- 
for ease of illustration.
The \CNN\ algorithm can be generalised freely to multiclass cases by considering pairwise
interactions between the propensity scores for each domain for applying the calipers and by
computing the pairwise inequalities between $\mbf{s}_q$ and $\mbf{s}_k$ (giving the connectivity
matrix $(\mbf{s}_q \cdot \mbf{1}^T) \neq (\mbf{1} \cdot \mbf{s}_k^T)$, where $\mbf{1}$ denotes the
ones vector of the same shape as its multiplicand and mediates broadcasting) for enforcing the
cross-domain constraint.
 
\section{Matching for PACS dataset}\label{appx:pacs_matching}
\begin{figure}[ht!]
     \centering
     \begin{subfigure}[b]{0.49\textwidth}
         \centering
         \includegraphics[scale=0.2]{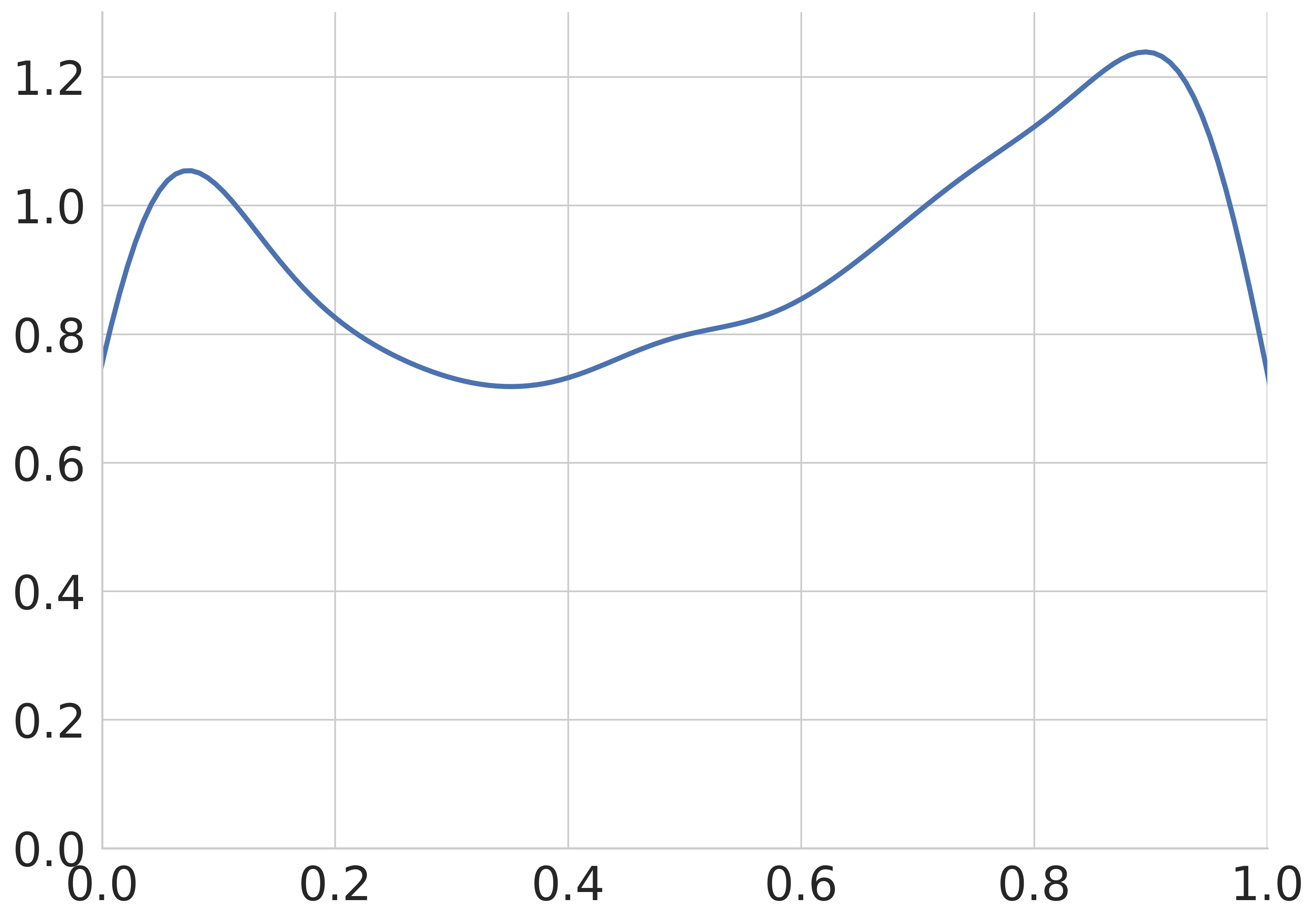}
         \caption{No temperature scaling ($\tau=1$).}
     \end{subfigure}
     \hfill
     \begin{subfigure}[b]{0.49\textwidth}
         \centering
         \includegraphics[scale=0.2]{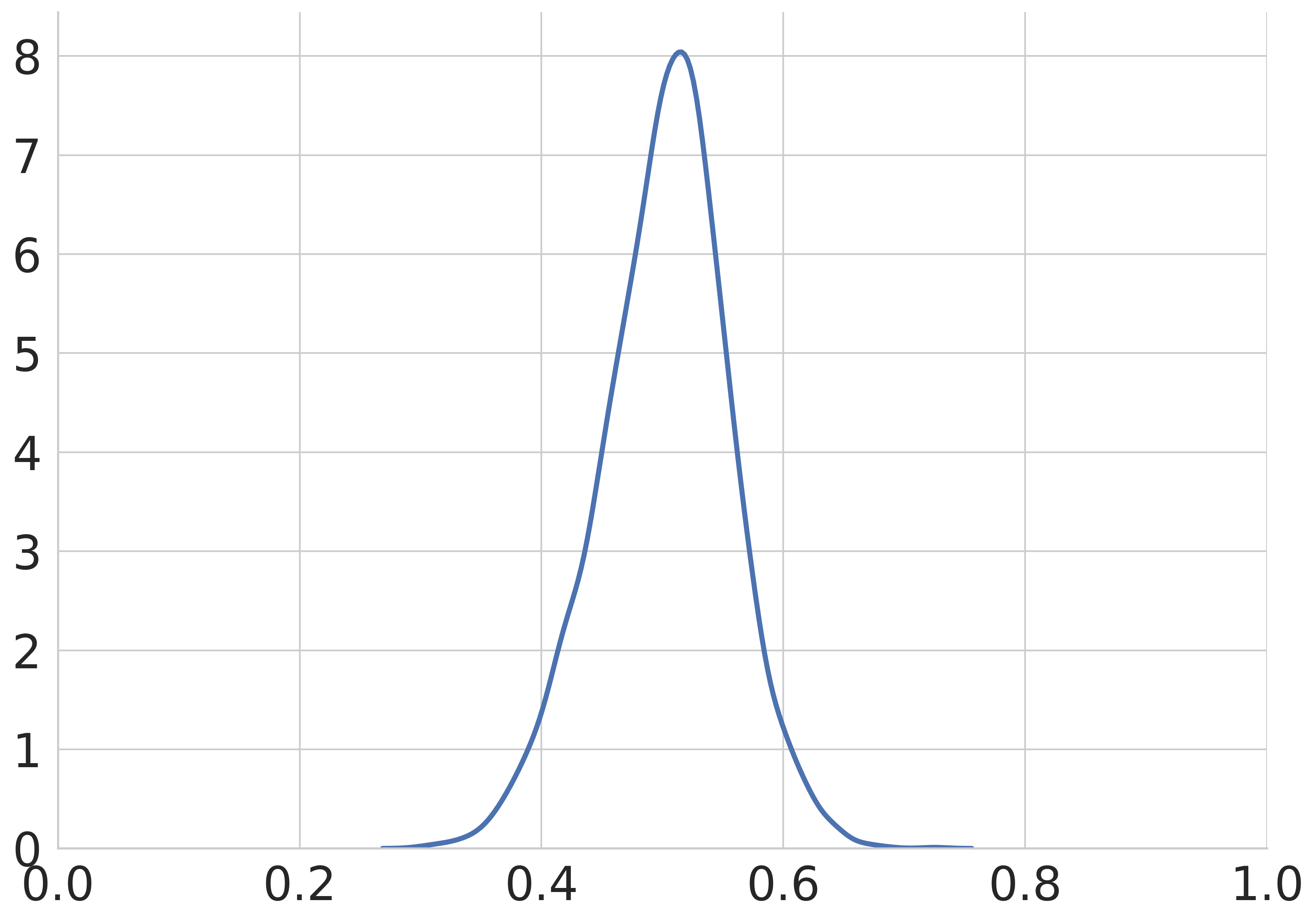}
         \caption{Temperature scaling, with $\tau=10$.}
     \end{subfigure}
    \caption{Estimated propensity score distribution of \textit{photo} and \textit{art painting} on the PACS dataset. We compare (a) the original distribution ($\tau=1$) and (b) the temperature-scaled distribution ($\tau=10$).
    Here, the large temperature has the effect of transforming a bimodal distribution into a unimodal one.}
    \label{fig:pacs_ps_pa}
\end{figure}

\begin{figure}[ht!]
     \centering
     \begin{subfigure}[b]{0.49\textwidth}
         \centering
         \includegraphics[scale=0.2]{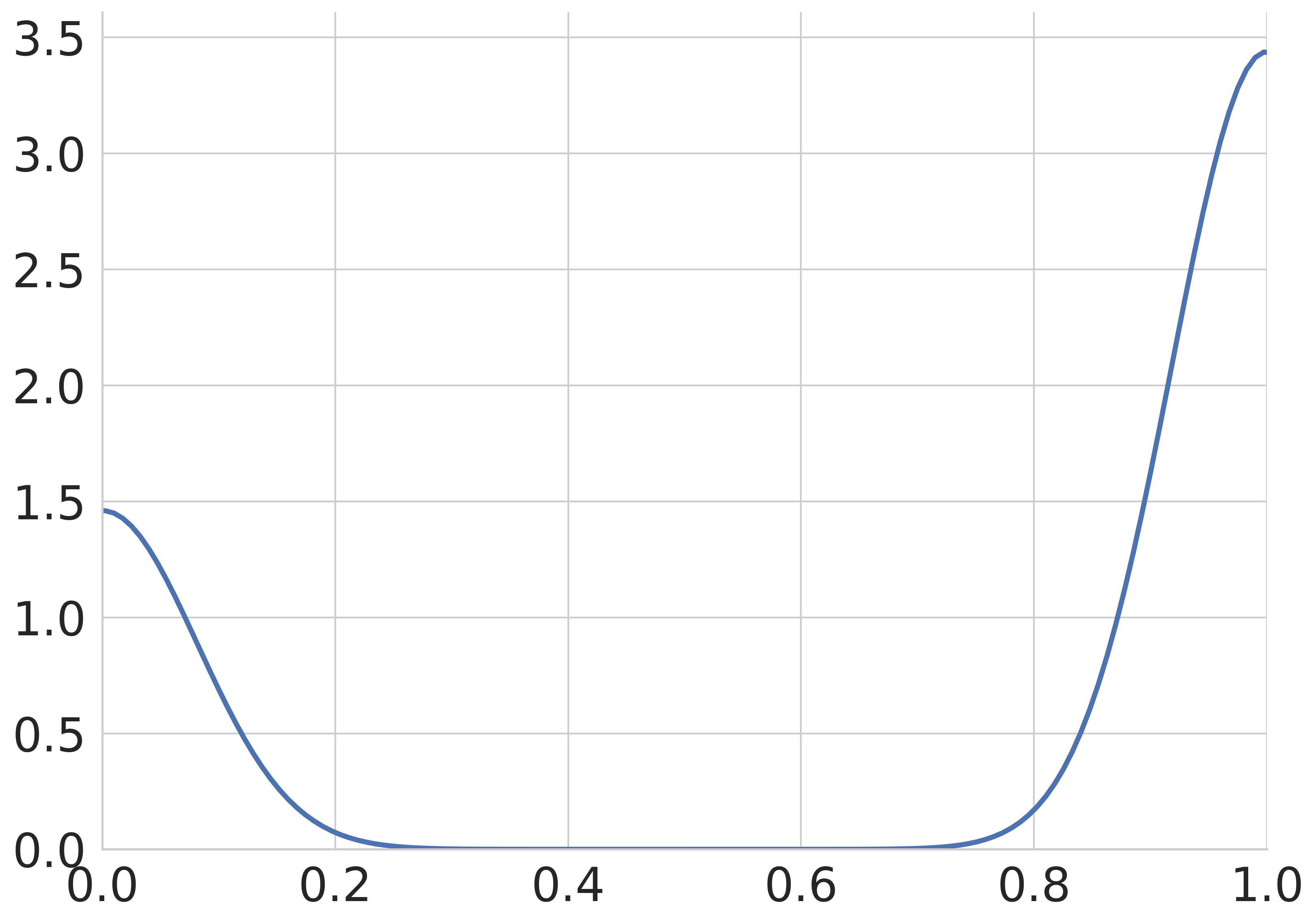}
         \caption{No temperature scaling ($\tau = 1$).}
     \end{subfigure}
     \hfill
     \begin{subfigure}[b]{0.49\textwidth}
         \centering
         \includegraphics[scale=0.2]{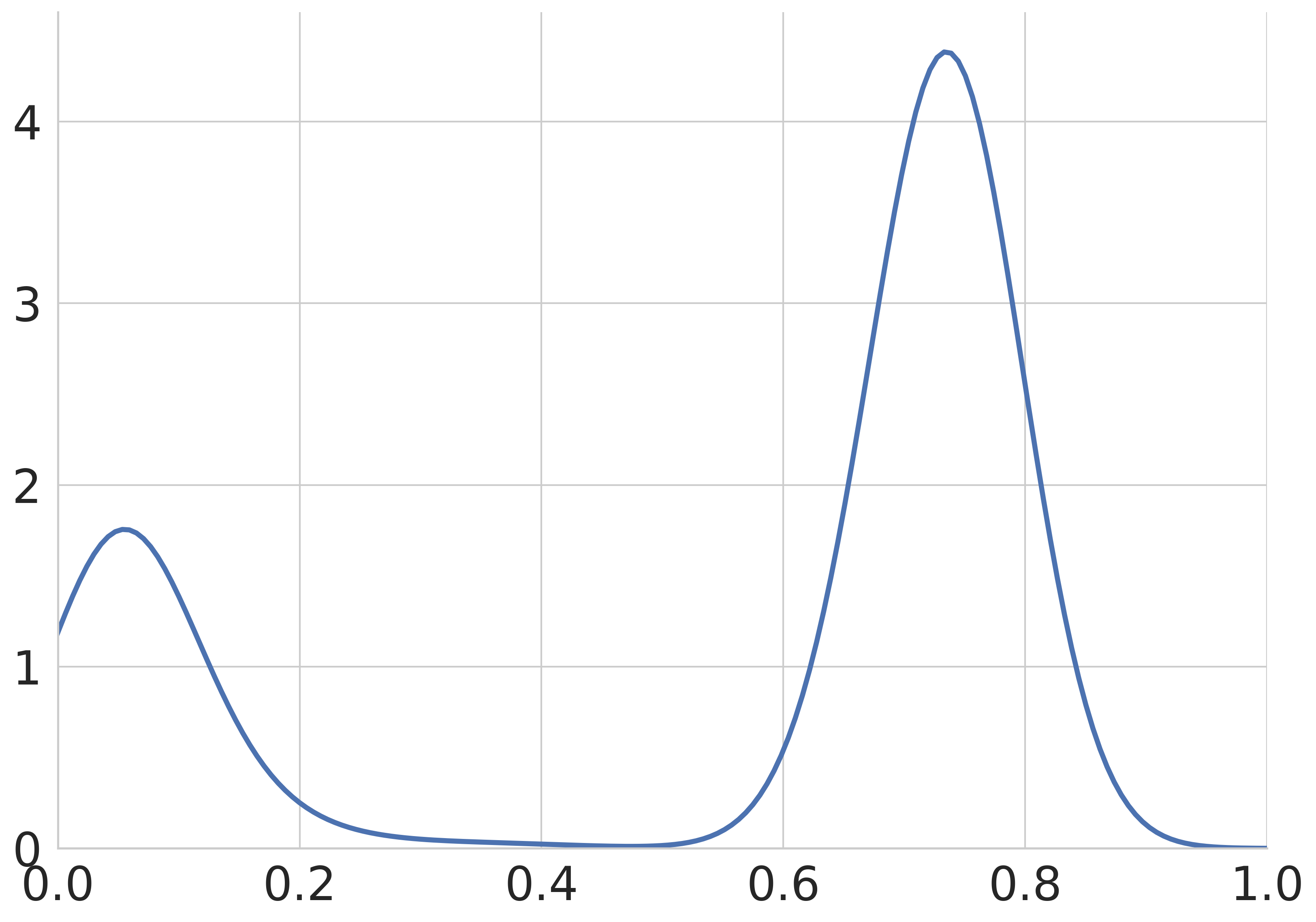}
         \caption{Temperature scaling, with $\tau=10$.}
     \end{subfigure}
    \caption{Estimated propensity score distribution of \textit{photo} and \textit{sketch} on the
    PACS dataset. We compare (a) the original distribution ($\tau=1$) and (b) the
temperature-scaled distribution ($\tau=10$). Here, the large temperature has the effect of smoothing the distribution.}
    \label{fig:pacs_ps_ps}
\end{figure}

In this section we perform some initial experiments on PACS dataset~\cite{li2017deeper} (using features extracted for a
pre-trained CLIP \citep{radford2021learning} model) to show how the temperature scaling can be used
to smooth the propensity score distribution to better control how many sample are discarded during
matching. 
There are 1,670 \textit{photo}, 2,048 \textit{art painting}, 2,344 \textit{cartoon}, and 3,929 \textit{sketch} in the
dataset. Here will evaluate the results of matching across the two domains \textit{photo} and
\textit{art painting} as well as across \textit{photo} and \textit{sketch}.
In Fig.~\ref{fig:pacs_ps_pa} and Fig.~\ref{fig:pacs_ps_ps} we compare the shape of the estimated
propensity score with its scaled version using a temperature value of 10. 
As we can see, in the case of a distribution with extremely heavy tails (photo, sketch), the effect
of smoothing the distribution is that when a fixed caliper is applied most of the samples are
retained. 
On the other hand, when the initial distribution is smoother, a temperature of 10 is extreme,
having the effect of transforming the bimodal distribution to a unimodal one.
Additionally, we tabulate in Table~\ref{tab:pacs_nsamples} the number of matched pairs retrieved when matching across the two domains photo and sketch; here we can see that by increasing the temperature we smooth the estimated propensity score distribution and thereby retain more samples. Similarly, we can retrieve more pairs by reducing the fixed caliper threshold. We also analyse the case of matching across the two domains photo and art painting. Using a fixed caliper defined defined by a threshold $t_f = 0.1 $ and no temperature scaling (i.e. $\tau=1$) the algorithm retrieves 1,142 pairs matching in the direction photo $\rightarrow$ art and 1,501 in the direction art $\rightarrow$ photo. 

\begin{figure}[ht!]
  \centering
  \includegraphics[width=1.\textwidth]{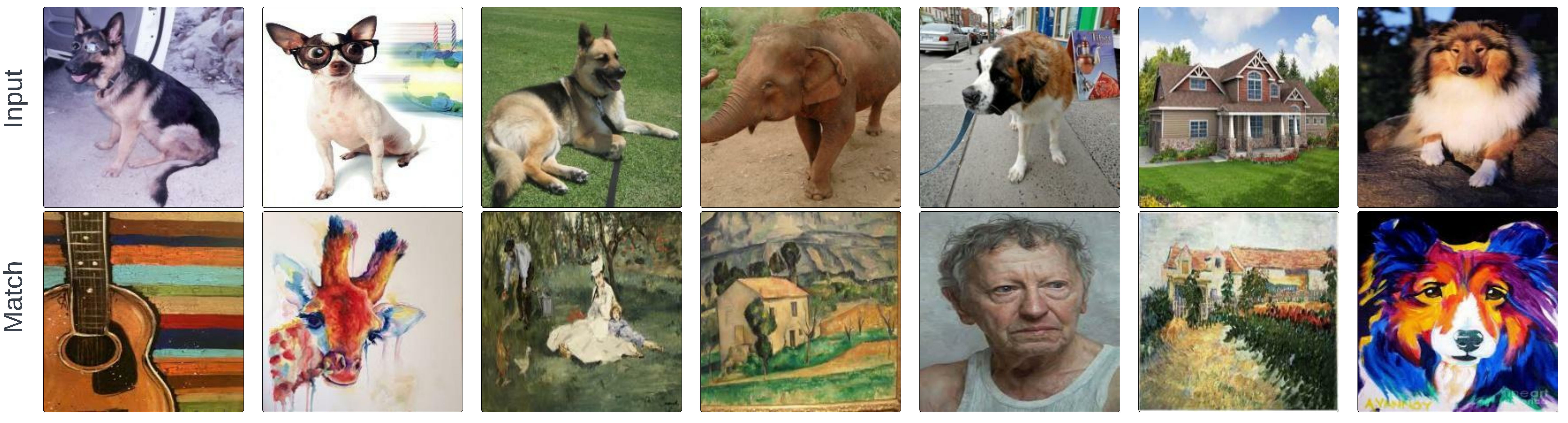}
  \caption{Examples of input (photo) images and their 1-NN matched (art paint) images retrieved using \CNN\ from the PACS dataset. 
  }
  \label{fig:match_pairs_pa}
\end{figure}

\begin{figure}[ht!]
  \centering
  \includegraphics[width=1.\textwidth]{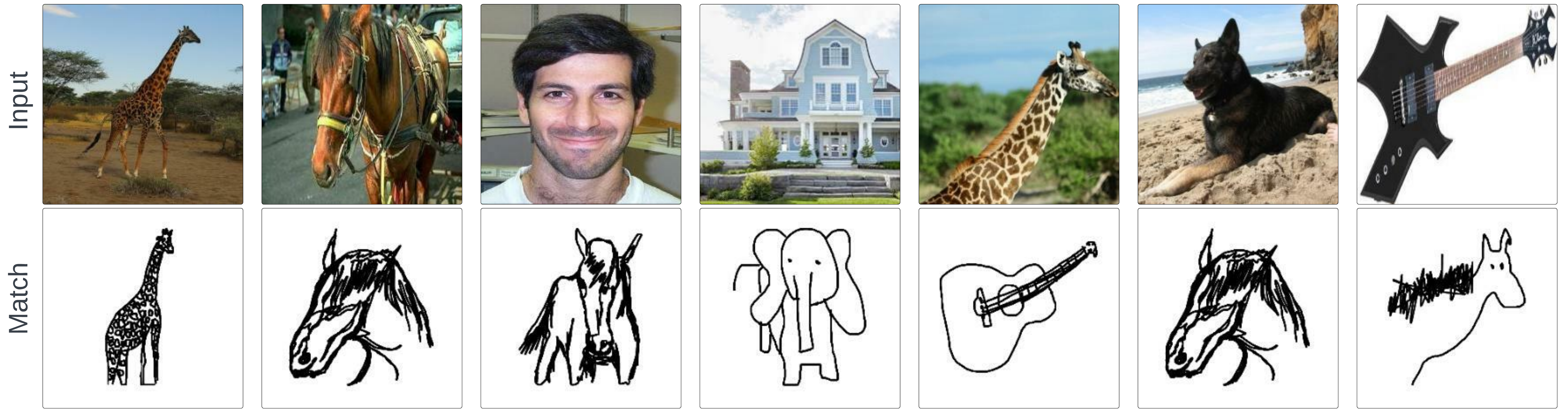}
  \caption{Examples of input (photo) images and their 1-NN matched (sketch) images retrieved using \CNN\ from the PACS dataset.
  }
  \label{fig:match_pairs_ps}
\end{figure}

In Fig.~\ref{fig:match_pairs_pa} and Fig.~\ref{fig:match_pairs_ps} we show examples of matching
pairs found using our \CNN\ algorithm. Although the features were not fine-tuned on PACS, we can see
a few examples of intraclass matching. For the photo-art painting application we can see
preservation in colour and background; while in the photo-sketch case shape and pose.


\begin{table}[ht!]
	\centering
	\caption{Analysis of the number of the retrieved matched pairs when matching across the two domain \textit{photo} and \textit{sketch} on the PACS dataset. The fixed caliper threshold and temperature scaling can be used to smooth the propensity score distribution and effect the number of pairs.}
	\scalebox{1.0}{
	\begin{tabular}{llll} 
	\toprule 
    \textbf{Fixed Caliper ($t_f$)}  & \textbf{Temperature ($\tau$)} & \textbf{photo} $\rightarrow$ \textbf{sketch} & \textbf{sketch} $\rightarrow$ \textbf{photo} \\
    \midrule
    0.1 & 1 & 0 & 0 \\
    0 & 1 & 1540 & 3929 \\
    0.01 & 1 & 6 & 9 \\
    0.01 & 1.3 & 14 & 56 \\
    0.01 & 1.8 & 25 & 574 \\
    0.01 & 2.5 & 41 & 3082 \\
    0.01 & 10 & 1540 & 3929 \\
    0.1 & 10 & 298 & 3929 \\
	\bottomrule
	\end{tabular}}
	\label{tab:pacs_nsamples}
\end{table}

\section{Energy and Carbon Footprint Estimates}\label{appx:carbon}
\import{./}{carbon.tex}

%% file: datasets.tex
We evaluate Okapi using three datasets -- iWildCam, PovertyMap, and CivilComments -- taken from the
WILDS 2.0 benchmark
\citep{SagWeiLeeGaoetal22}.
These datasets were chosen specifically due to the poor performance reported by
\cite{SagWeiLeeGaoetal22} for semi-supervised and domain adaptation methods across the board, in
relation to the ERM baselines.
For PovertyMap in particular, ERM was found to vastly outperform any competing methods utilising
the unlabelled data and/or domain labels.

\textbf{iWildCam-WILDS} is an extension of the iWildCam 2020 Competition Dataset
\cite{beery2020iwildcam}. 
The task is multi-class species classification  of animals in camera trap images. The dataset
contains 1022K images of animals annotated with the domain, $s$, that identifies the camera trap
that captured it. The target label, $y$, is one of 182 different animal species and it is provided
solely for the 203K labelled data. The labelled training set contains 130K images taken by 243 camera traps. 
The out-of-distribution (OOD) validation and target sets include images from 32 and 48 different
camera traps which are disjoint from the 243 training domains.
Additionally, 819K unlabelled images from 3215 new domains are available. 
Different cameras trap differ in characteristics such as illumination, background and relative
animal frequency, models trained on the source domains might fail to generalise to images taken
from new locations.

\textbf{PovertyMap-WILDS} is a variation of the dataset introduced in 
\cite{yeh2020using}. The task is to predict the wealth index, $y$, from multispectral satellite
images of 23 African countries. The country the image was taken in as well as whether it was taken
in a rural or urban area represent the domain $s$. The dataset contains 5 cross-validation (CV)
folds of roughly equal size, each one dividing the 23 countries differently across the source, OOD
validation and OOD target splits. In each fold, the labelled training set contains 11K images from
14 different countries. The OOD validation and target sets include images from 5 different
countries not represented in the source data.
The dataset also includes 261K unlabelled images from the same 23 countries. 

\textbf{CivilComments-WILDS} is an online-comment dataset adapted from
\cite{borkan2019nuanced},  comprising 448K online comments
annotated with both a binary indicator of toxicity ($\{\mathrm{\emph{toxic}, \emph{not toxic}}\}$)
-- serving as the target label, $y$ -- and the demographic identities mentioned within them --
serving as the
domain $s$. 
Here, $s \in \{0, 1\}^8$ is a binary vector rather than a scalar, with dimensions 
indicating membership (non-exclusively) to 8 demographic groups, spanning different genders,
religions and ethnicities.
For the WILDS 2.0 variant of the dataset, \cite{SagWeiLeeGaoetal22} introduce an additional corpus
of 1551K
comments acting as the unlabelled training data belonging to extra domains.
While the comments are completely unlabelled, w.r.t. both $y$ and $s$ and thus are not
domain-separable at the sample level, the majority ($92\%$) of the comments are known to be sourced
from the same documents as those comments comprising the (OOD) labelled test data. 
As noted in \cite{SagWeiLeeGaoetal22}, CivilComments-WILDS exhibits label imbalance w.r.t. $y$;
this is amended both therein and herein (as appertains all methods) through the use of
class-balanced sampling, though with the minor distinction that for our experiments we ensure each
batch is exactly balanced rather by sampling equally from each class, in contrast to
\cite{SagWeiLeeGaoetal22} who sample hierarchically -- sampling $y_i$ uniformly from $\{0, \dots,
|\mc{Y}_l|\}$ and then uniformly from $\mc{D}_l$, conditioned on $y_i$ -- such that balance is
achieved only in expectation.

%% file: fairness.tex
DG and Algorithmic Fairness overlap in their objective to train a model that yields predictions that are
statistically independent of (and thus robust to variations in) domain, when for the latter the
domain is taken to be some protected characteristic, such as age or gender, and fairness is
measured according to invariance-driven notions of group fairness such as Demographic Parity
\citep{feldman2015certifying} and Equal Opportunity \citep{hardt2016equality}. Indeed, methods that
focus on equalising the empirical risk across subgroups -- such as by importance weighting
\citep{idrissi2022simple, shimodaira2000improving} -- have featured extensively in both DG
\citep{arjovsky2019invariant,creager2021environment,krueger2021out,sagawa2019distributionally} and
fairness \citep{agarwal2018reductions,donini2018empirical,kamiran2012data} and many approaches to
fair representation learning \citep{ creager2019flexibly,kehrenberg2020null,madras2018learning,
oneto2020exploiting,quadrianto2019discovering} have roots in the former \cite{muandet2013domain}
and in the closely-related field of domain adaptation \citep{ganin2016domain}. Beyond this more
general equivalence, our work also has ties to notions of individual fairness pioneered by
\cite{dwork2012fairness} -- broadly prescribing that similar individuals be treated similarly -- in
that our unsupervised loss involves maximising the similarity between inter-domain samples within
representation space. This is reminiscent of the operationalisation of individual fairness proposed
by \cite{lahoti2019operationalizing} that enforces similarity between a given representation and
the representations of its neighbouring -- in both the input space and according to a between-group
(cross-domain) quantile graph -- samples.

%% file: ext_results.tex
\begin{table}[tbp]
	\centering
	\caption{
	  An extended comparison between Okapi and different baselines on two benchmark image
	  datasets. We include both the results of our re-run of the baselines and those of
	  \cite{SagWeiLeeGaoetal22}. Both ID and OOD performances are reported. For iWildCam we
	  average over results from 3 different seeds, for PovertyMap we do so over the 5
	  pre-defined CV folds.
	  Standard deviations are shown in parentheses.
	  The additions relative to Table~\ref{tab:iw_pm_results} include results with an offline
	  variant of Okapi -- where the matches are generated prior to training from the features
	  of the trained ERM model and then fixed for the course of training -- and with the ResNet
	  backbones employed by \cite{SagWeiLeeGaoetal22}.
      }
	\scalebox{0.75}{
	\begin{tabular}{lllllll}
	\toprule \textbf{Method} & \multicolumn{2}{c}{\textbf{iWildCam}} &
	\multicolumn{4}{c}{\textbf{PovertyMap}} \\ & \multicolumn{2}{c}{macro F1 $\uparrow$} &
	\multicolumn{2}{c}{worst U/R corr. $\uparrow$} & \multicolumn{2}{c}{worst U/R MSE
	$\downarrow$} \\
	\midrule
        & ID & OOD & ID & OOD & ID & OOD \\
        ERM \cite{SagWeiLeeGaoetal22} & 47.0 (1.4) & 32.2 (1.2) & 0.66 (0.04) & 0.49 (0.06) & - & - \\
        FixMatch \cite{SagWeiLeeGaoetal22} & 46.3 (0.50) & 31.0 (1.3) & 0.54 (0.10) & 0.30 (0.11) & - & - \\
	\midrule
        ERM (ConvNeXt) & 48.6 (1.1) & 33.3 (0.3) & 0.72 (0.03) & 0.53 (0.09) &  0.23 (0.03) & 0.35 (0.12)\\
	%
	%
        FixMatch (ConvNeXt) & 51.1 (1.0) & 35.2 (0.7) & 0.50 (0.13) & 0.34 (0.12) & 0.59 (0.42) & 0.88 (0.61)\\
        Okapi (ours; ConvNeXt) & 50.6 (0.7) & 36.1 (0.9) & 0.72 (0.02) & 0.55 (0.10) & 0.22 (0.02) & 0.33 (0.10)\\
	Okapi (offline; ConvNeXt) & 48.8 (0.8) & 31.7 (0.2) & 0.68 (0.02) & 0.53 (0.07) & 0.26 (0.02) & 0.37 (0.13) \\
        Okapi (no calipers; ConvNeXt) & - & - & 0.72 (0.02) & 0.54 (0.12) & 0.22 (0.02) & 0.36 (0.14) \\
	\midrule
        ERM (ResNet) & 46.5 (0.8) & 29.7 (1.0) & 0.69 (0.03) & 0.53 (0.08) &  0.24 (0.04) & 0.34
	(0.11)\\
        FixMatch (ResNet) & 43.0 (2.5) & 25.5 (1.4) & 0.70 (0.02) & 0.53 (0.08) &  0.24 (0.02) &
	0.35 (0.10)\\
        Okapi (ResNet) & 46.1 (0.7) & 27.8 (0.3) & 0.70 (0.04) & 0.52 (0.07) &  0.23 (0.02) &
	0.33 (0.10)\\
	\bottomrule
	\end{tabular}}
	\label{tab:ext_iw_pm_results}
\end{table}

%% file: implementation.tex
\paragraph{Data Augmentation} We follow \cite{SagWeiLeeGaoetal22} when defining the augmentations
for the the WILDS datasets.
In the case of PovertyMap-WILDS we corroborate the original finding that data-augmentation
adversely affects performance, and, in light of this, elect only to use data-augmentation for
FixMatch where it is needed to generate the weak and strong views used in computing the consistency
loss.
Since Okapi uses an NN-based approach for generating these views, it is decoupled from the
augmentation strategy and problems that can arise from its misspecification.

\paragraph{Architecture} For our image experiments, contrary to \cite{SagWeiLeeGaoetal22}, we
opt to use the recently proposed ConvNeXt architecture \citep{liu2022convnet}, finding this change
to provide large performance gains and to be crucial in enabling semi-supervised methods to surpass
the ERM baseline.
This is in line with \cite{kim2022broad} who similarly found that a change of architecture
(combined with large-scale pre-training) could greatly bolster performance on the iWildCam dataset.
More precisely, we use the \emph{tiny} variant of ConvNeXt, pre-trained on ImageNet 1k, as the
initial backbone for our models. We compose this with a single fully-connected layer to construct
the complete predictor both for the target and the propensity score. For our CivilComments
experiments, in contrast, we do not diverge from \cite{SagWeiLeeGaoetal22} in our choice of
architecture, with all models trained with a pre-trained DistilBERT \citep{sanh2019distilbert}.

\paragraph{Optimisation} For
optimising all models, we use the AdamW optimiser \citep{loshchilov2018decoupled} coupled with a
cosine annealing schedule without warm restarts \citep{DBLP:conf/iclr/LoshchilovH17}.
We set the initial learning to be $1 \times 10^{-4}$ across the board, and forgo the use of weight
decay. 
Models are trained for $120$K, $30$K, and $20$K iterations for iWildCam, PovertyMap, and
CivilComments, respectively.
 The decay coefficient, $\zeta$, for the target encoder's exponential moving average is initialised
 to $\zeta_{\mr{start}}$ and is linearly increased to $\zeta_{\mr{end}}$ over the course of
 training. For PovertyMap we set $\zeta_{\mr{start}}$ and $\zeta_{\mr{end}}$ to be $0.996$ and
 $1.0$, respectively;
 for iWildCam, we set $\zeta_{\mr{start}}$ and $\zeta_{\mr{end}}$ to be
 $0.999$ and $0.999$, respectively, resulting in a fixed value of $\zeta$; 
 $1.0$, respectively; 
 for CivilComments, we set $\zeta_{\mr{start}}$ and $\zeta_{\mr{end}}$ to be
 $0.996$ and $0.996$, respectively, again resulting in a fixed value of $\zeta$. 
We similarly warm up the pre-factor for the consistency loss, $\lambda$, according to a linear
schedule during the first $10\%$ of training to allow a period for the encoder to learn meaningful
relations between samples through the supervised loss before bootstrapping with the consistency
loss, with a final value of $1$.

\paragraph{Matching}\label{matching_imp} In order to determine suitable hyperparameters, $\xi$, for
\CNN, we perform a grid-search in the static setting, using a fixed model. Specifically, we use the
backbone of an ERM-trained model as the encoder with which to generate the queries and keys for
matching.
The quality of matching with a given instantiation of $\xi$ is measured using two metrics commonly
used in the statistical matching literature: \emph{Variance Ratio} (VR) and \emph{Standard Mean
Differences} (SMD) \citep{rubin2001using}. 
Both metrics operate on pairs of domains, but can be generalised to work when $s$ is
non-binary by simply aggregating over all pairwise results.
For a given pair of domains, VR is defined as the ratio of the variances of the covariates between
the two domains, with an ideal value of $1$, while SMD is defined as the difference in their
covariate means, normalised by the standard deviation for each covariate, and is to be minimised.
While our proposed method is applicable whether $S$ is binary or categorical, for the experiments
in this paper we take advantage of the fact that the WILDS datasets specify splits with
non-overlapping domains and match from $\D_l \rightarrow \D_u$ and in the reverse direction (from
$\D_u \rightarrow \D_l$). This decision was based on preliminary experiments which found the binary
variant generally enjoyed more stable optimisation, something which future work should seek to
rectify.
In the case of PovertyMap, however, the training splits themselves do not satisfy the
aforementioned requirement of being sourced from mutually exclusive sets of domains and we instead
treat the OOD validation set as $\D_u$ (and treat it as being unlabelled w.r.t. $y$, in that it is
only used for $\mc{L}_\mr{unsup}$).

%% file: tables/pm_ablations.tex
\begin{table}[ht!]
	\centering
 	\caption{Ablation experiments for Okapi conducted using the PovertyMap-WILDS dataset. Specifically, we assess the importance of four elements of our proposed method: the number of nearest neighbours used in computing $\mc{L}_\mr{unsup}$ ($k$), the use (enabled/disabled) of the fixed- and std-calipers in $\mr{\CNN}$ (considering these to be separate components), and which encoder (online or target) is used to generate the queries for statistical matching (with use of the target encoder `TE queries' being the default and `OE queries' denoting the alternative). Both ID and OOD performances are reported. The results are computed by aggregating over the results for each of the 5 pre-defined cross-validation folds. We report the average and standard deviation value across replicates of the metric of interest.}
	\begin{subtable}[t]{\textwidth}
    	\centering
    	\caption{$\mr{\CNN}$ ablations.}
    	\scalebox{1}{
    	\begin{tabular}{lllll}
    		\toprule
    		\textbf{Method} & \multicolumn{4}{c}{\textbf{PovertyMap}} \\
    		& \multicolumn{2}{c}{worst U/R corr. $\uparrow$} & \multicolumn{2}{c}{worst U/R MSE $\downarrow$} \\
    		\midrule
            & ID & OOD & ID & OOD \\
            Okapi (k=1) & 0.72 (0.02) & 0.55 (0.10) & 0.22 (0.02) & 0.33 (0.10) \\
            Okapi (k=5) & 0.72 (0.02) & 0.55 (0.10) & 0.22 (0.02) & 0.33 (0.10)\\
            Okapi (k=10) & 0.72 (0.02) & 0.55 (0.09) & 0.22 (0.02) & 0.33 (0.10)\\
            Okapi (k=5, no calipers) & 0.72 (0.02) & 0.54 (0.12) & 0.22 (0.02) & 0.36 (0.14) \\
            Okapi (k=5, no std caliper) & 0.72 (0.02) & 0.54 (0.12) & 0.22 (0.02) & 0.35 (0.14) \\
            Okapi (k=5, no fixed caliper) & 0.72 (0.02) & 0.55 (0.10) & 0.22 (0.02) & 0.33 (0.10) \\
    		\bottomrule
    	\end{tabular}}
	\end{subtable}
	
	\vspace{5mm}
	
	\begin{subtable}[t]{\textwidth}
    	\centering
    	\caption{Query-generator ablation (target encoder (TE) vs. online encoder (OE)).}
    	\scalebox{1}{
    	\begin{tabular}{lllll}
    		\toprule
    		\textbf{Method} & \multicolumn{4}{c}{\textbf{PovertyMap}} \\
    		& \multicolumn{2}{c}{worst U/R corr. $\uparrow$} & \multicolumn{2}{c}{worst U/R MSE $\downarrow$} \\
    		\midrule
            & ID & OOD & ID & OOD \\
            Okapi (TE queries) & 0.72 (0.02) & 0.55 (0.10) & 0.22 (0.02) & 0.33 (0.10)\\
            Okapi (OE queries) & 0.72 (0.02) & 0.55 (0.10) & 0.23 (0.02) & 0.34 (0.10) \\
    		\bottomrule
    	\end{tabular}}
	\end{subtable}
	
	 \label{tab:ablations}
\end{table}

%% file: pseudocode/calipernn.tex
\begin{algorithm}[ht]
     \caption{
         Pytorch-style pseudocode for the \CNN\ matching algorithm for the special case where the
         domain is binary. The algorithm generalises freely to arbitrary numbers of domains however
         we restrict ourselves to the binary version here for illustrative purposes.
}
    \label{alg:calipernn_pc}
    \begin{minted}{python}

    def binary_caliper_nn(
        x_query, # samples to be used as the queries for matching
        s_query, # binary labels indicating the domain of x_query
        x_key, # samples to which the query samples may be matched to.
        s_key, # binary labels indicating the domain of x_key
        ps_query, # propensity scores associated with x_query
        ps_key, # propensity scores associated with x_key
        t_f, # threshold for the fixed caliper
        t_sigma, # number of standard deviations at which to threshold
        k # number of neighbours to attempt to retrieve per query
    ):
        anchor_inds, positive_inds = [], []
        for direction in (0, 1): # which domain (0 or 1) to treat as the 'anchor'
            key_mask = s_key != direction
            # exclude samples with propensity scores outside the valid range
            # determined by t_f: (1 - t_f, t_f)
            fc_mask = (ps_query > (1 - t_f)) &  (ps_query < t_f)
            anchor_mask = fc_mask & (s_query == direction)
            queries_x_filtered = queries.x[anchor_mask]
            ps_query_filtered = ps_query[anchor_mask]
            fc_mask = (ps_key > (1 - t_f)) & (ps_key < t_f)
            key_mask &= fc_mask
            ps_key_filtered = ps_key[key_mask]
            # 2-norm distance between unfiltered propensity scores
            dists_ps = cdist(ps_query_filtered, ps_key_filtered, p=2) 
            # 2-norm distance between the filtered anchors and keys
            dists_x = cdist(queries_x_filtered, x_key[key_mask], p=2)
            # compute sigma as the mean of the per-domain standard deviations
            std_ps = (0.5 * (ps_query_filtered.var() + ps_key_filtered.var())).sqrt() 
            std_threshold = t_sigma * std_ps
            # filter out any samples that violate the std-caliper
            dists_x[dists_ps > std_threshold] = float("inf")
            nbr_dists, nbr_inds = dists_x.topk(dim=1, largest=False, k=k)
            # filter out queries not yielding the requisite number of matches (k)
            is_matched = ~nbr_dists.isinf().any(dim=1)

            anchor_inds.append(anchor_mask.nonzero()[is_matched])
            positive_inds.append(key_mask.nonzero()[nbr_inds[is_matched]])

        return cat(anchor_inds, dim=0), cat(positive_inds, dim=0)

    \end{minted}
\end{algorithm}

%% file: pseudocode/ol.tex
\begin{algorithm}[ht]
     \caption{
         Pytorch-style pseudocode for the online learning algorithm for the special case where the
         labelled and unlabelled datasets are treated as the domains. The algorithm generalises
         freely to arbitrary numbers of domains however we restrict ourselves to the binary version
         here for illustrative purposes.
    }
    \label{alg:ol_pc}
    \begin{minted}{python}
    # online_encoder: online encoder
    # predictor_head: online predictor head
    # propensity_scorer: online propensity scorer
    # target_encoder momentum encoder (frozen)
    # n_m: memory-bank capacity
    # zeta: decay rate of the EMA updates
    # tau: temperature-scaling parameter for the propensity scores.
    # t_f: fixed caliper threshold for CaliperNN
    # t_sigma: number of standard deviations at which to threshold in CaliperNN
    # l_sup: supervised loss function
    # k: number of matches to retrieve per query
    # lambda_: loss pre-factor for the unsupervised loss
    # D: Dimensionality of the encodings.

    feature_mb = empty(n_m, D) # memory bank storing momentum-encoded features
    label_mb = empty(n_m) # memory bank storing domain labels associated with feature_mb
     # load minibatches with B_l labelled samples and B_u unlabelled samples
    for x_l, y, x_u in train_loader:
        # EMA update: \theta^\prime_t = \zeta \theta^\prime_{t - 1} + (1 - \zeta) \theta_t
        ema_update(target_encoder, online_encoder, zeta)
        features_o_l = online_encoder(x_l) # f_\theta(x_l) -> z_l
        features_t = target_encoder(cat([x_l, x_u])) # f_\theta(x_l \cup x_u) -> z_q^\prime
        y_hat = predictor_head(features_o_l) # g_\phi(z_l) -> \hat{y}
        features_o_u = online_encoder(x_u) # f(x_u) -> z_u
        features_o = cat([features_o_l, features_o_u]) # z_q := z_l \cup z_u
        # normalize the encodings to unit vectors.
        features_o_n = normalize(features_o, p=2, dim=1)
        queries = normalize(features_t, p=2, dim=1)
        # we treat x_l and x_u as coming from domains indexed by 0 and 1, respectively
        labels_l_q = ones(len(x_l)) # ones-vector of size B_l
        labels_u_q = zeros(len(x_u)) # zeros-vector of size B_u
        labels_q = cat([labels_l_q, labels_u_q])
        mb_mask = is_empty(label_mb) # mask indicating which elements of the MB are filled
        labels_k = cat([labels_q, label_mb[mb_mask].clone()])
        # keys are the union of the queries and the memory-bank-stored features
        keys = cat((queries, feature_mb[mb_mask].clone()), dim=0)
        feature_mb.push(queries) # update the feature memory bank
        label_mb.push(labels_q) # update the label memory bank
        logits_ps_k = propensity_scorer(keys) # h_\psi(z_k) -> e_k
        loss_ps = xent(logits_ps_k, labels_k) # (binary) cross-entropy loss
        # tempered logistic function: 1 / (1 + exp(-logits_ps_k / tau))
        logits_ps_k = sigmoid(logits_ps_k / tau) 
        logits_ps_q = logits_ps_k[:len(queries)]
        # filter and match queries with (binary) CaliperNN
        inds_a, inds_p = binary_caliper_nn(
            features_t_n, labels_q, keys, labels_k,
            logits_ps_q, logits_ps_k, t_f, t_sigma, k
        )
        # compute the unsupervised loss (d(z_q, z_n)) for all matched queries
        z_q, v_k = features_o[inds_a], keys[inds_p]
        match_rate = len(z_q) / len(features_o)
        loss_u = match_rate * (z_q.unsqueeze(1) - v_q).pow(2).sum(-1).mean()
        loss = l_sup(y_hat, y) + lambda_ * loss_u + loss_ps # aggregate loss
        loss.backward() # compute gradients
        update(online_encoder, predictor_head, propensity_scorer) # optimizer updates

    \end{minted}
\end{algorithm}

%% file: carbon.tex
To highlight the efficiency of Okapi, we provide estimates in \ref{tab:carbon_fp} of the carbon footprint associated
with the running of it and of the ERM and FixMatch baselines on the iWildCam dataset, using the
same hyperparameter configuration used to generate the results in the main text.
The runs were conducted in a controlled fashion, using the computing infrastructure and device
count in all cases.

\begin{table}[htp]
	\centering
	\caption{
	  Comparison of the estimated carbon footprint (kgCoeq) of Okapi with the ERM and
	  FixMatch baselines per replicate of the iWildCam dataset. 
	  For the controlled training conducted to enable fair computation of these estimates, we
	  used a private infrastructure with an estimated
	  carbon efficiency of 0.432 kgCOeq/kWh and RTX 3090 GPUs, each job being run on a single
	  GPU, coupled with four data-loading workers.
	}
	\scalebox{1.0}{
	\begin{tabular}{lc} 
	\toprule 
	\textbf{Method} & \textbf{kgCOeq} $\downarrow$ \\ 
	\midrule
        ERM & 1.36 \\
	FixMatch & 2.12 \\
	Okapi (ours) & 1.97 \\
	\bottomrule
	\end{tabular}}
	\label{tab:carbon_fp}
\end{table}